%% file: submission.tex
\documentclass{article}

\PassOptionsToPackage{numbers,compress}{natbib}

\usepackage[preprint]{neurips_2026}

\usepackage[utf8]{inputenc} 
\usepackage[T1]{fontenc}    
\usepackage{hyperref}       
\usepackage{url}            
\usepackage{booktabs}       
\usepackage{amsfonts}       
\usepackage{nicefrac}       
\usepackage{microtype}      
\usepackage{xcolor}         
\usepackage{wrapfig}
\usepackage{paracol}

\def\eg{\emph{e.g.}}
\def\ie{\emph{i.e.}}

\def\name{LangTail}

\usepackage{multirow}
\usepackage{amsmath}
\usepackage{graphicx}
\usepackage{algorithm}
\usepackage{algorithmic}
\input{math_commands.tex}

\title{Resolving Long-Tail Ambiguity in Unsupervised 3D Point Cloud Segmentation with Language Priors}

\author{%
 \small
  Siqi Wei$^{1,}$\thanks{Equal contribution. $^\dagger$Corresponding authors.} \quad 
  Hongbin Xu$^{2,*}$ \quad 
  Feng Xiao$^{1}$ \quad 
  Tian Lan$^{3}$ \quad 
  Chun Li$^{4}$ \quad 
  Ming Li$^{5,\dagger}$ \quad 
  Qiuxia Wu$^{1,\dagger}$ 
  \vspace{0.2cm}\\
  \normalfont \small
  $^{1}$South China University of Technology \quad 
  $^{2}$Bytedance \\
  \normalfont \small 
  $^{3}$Tsinghua University \quad 
  $^{4}$Shenzhen MSU-BIT University \quad 
  $^{5}$Guangming Laboratory
}

\begin{document}

\maketitle


\begin{abstract}

Existing approaches for unsupervised 3D point cloud segmentation predominantly rely on a \textit{purely visual} similarity-based learning-by-clustering paradigm, which suffers from a fundamental limitation: \textbf{long-tail ambiguity}. In such a paradigm, features of minor classes are consistently absorbed by dominant clusters, leading to severely imbalanced predictions.
To address this issue, we propose \textbf{\name{}}, a language-guided hierarchical learning framework that leverages the balanced world knowledge encoded in language models to mitigate long-tail ambiguity in unsupervised 3D segmentation. The key idea is to establish multi-level associations between language-derived semantic priors and visually underrepresented minor classes, thereby compensating for the biased attention of purely visual clustering toward dominant classes.
Specifically, \name{} first constructs an entity-level semantic prior from language models, capturing balanced and fine-grained world knowledge across categories. These priors are injected into a hierarchical clustering framework via contrastive alignment. This guides multi-granularity semantic structure formation and prevents minor classes from being absorbed by dominant clusters, yielding more discriminative representations for underrepresented categories.
Extensive experiments on ScanNet-v2, S3DIS, and nuScenes demonstrate that \name{} consistently outperforms existing methods by significant margins, \ie, +13.5, +12.9, and +8.9 mIoU, respectively. These results demonstrate the effectiveness of language priors in improving the representation of minority classes in 3D point clouds.
The code will be released at: \url{https://github.com/Whisky0129/langtail_official}.

\end{abstract}

\section{Introduction}

3D point cloud semantic segmentation is a fundamental task for a wide range of applications, including robotic planning \cite{krusi2017driving}, autonomous driving \cite{cui2021deep}, and embodied AI \cite{ze20243d,huang2026pointworld}. While recent fully supervised methods have achieved remarkable progress, they rely heavily on large-scale datasets with dense 3D annotations \cite{hu2020randla,wu2024point,kolodiazhnyi2024oneformer3d}, which are costly and labor-intensive to obtain. More importantly, in many real-world scenarios such as autonomous robotic systems, manual annotation is often impractical or even infeasible \cite{behley2019semantickitti,hu2021towards}.
To alleviate this dependency, a line of unsupervised methods has been proposed, including GrowSP \cite{zhang2023growsp}, PointDC \cite{chen2023pointdc}, and LogoSP \cite{zhang2025logosp}. These approaches typically follow a learning-by-clustering paradigm, where pseudo-labels are generated based on \textit{purely visual} features and iteratively refined to train the network.

\begin{figure}[tbp]
  \centering
  \includegraphics[width=\linewidth]{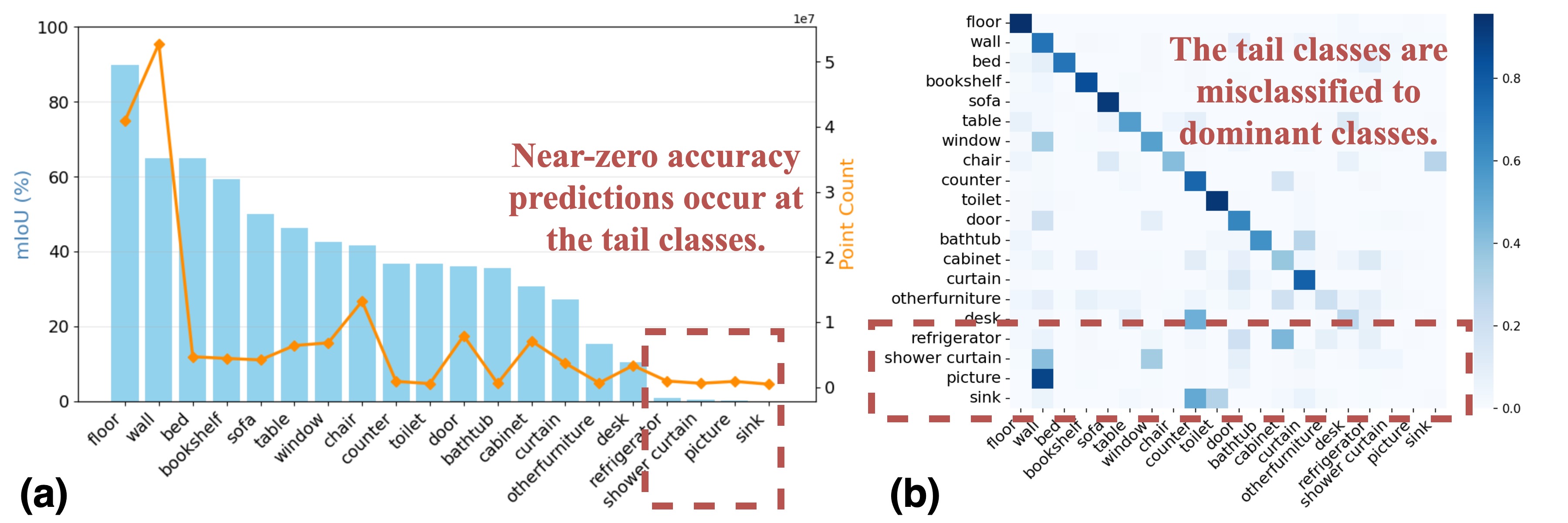}
  \vspace{-0.5cm}
  \caption{Analysis of the \textbf{Long-tail ambiguity} problem in unsupervised 3D semantic segmentation. (a) visualizes the per-category mIoU results (in \textcolor{cyan}{blue}) and the per-category point counts (in \textcolor{orange}{orange}). The long-tail amiguity problem can be observed, where near-zero predictions occur at the tail class. (b) visualizes the confusion matrix between the predicted results and the ground truth under each category. It can be observed that the tail classes are misclassified to dominant classes.}
  \label{fig:motivation}
  \vspace{-0.5cm}
\end{figure}

Despite their advances, existing methods suffer from a fundamental limitation: \textbf{long-tail ambiguity}. As shown in Fig.~\ref{fig:motivation}, while dominant classes are segmented reasonably well, tail classes are consistently misclassified and often achieve near-zero accuracy. This limitation arises because \textit{purely visual similarity-based clustering inherently favors majority patterns}, which appear more frequently in the data, causing features of rare categories to be absorbed into dominant clusters. As a result, the learned representations are biased toward frequent classes, leading to imbalanced and semantically inconsistent predictions.

In contrast, recent advances in vision-language models have shown that language models trained on large-scale corpora encode rich and structured world knowledge \cite{guo2025seed1,seed2026seed1,yang2025qwen3,bai2025qwen3}. Unlike purely visual representations learned from a limited point cloud dataset, such knowledge captures diverse semantic concepts and their relationships across a wide range of categories \cite{radford2021learning}. For example, language models can distinguish fine-grained semantic entities (\eg, \textit{shower curtain} vs. \textit{curtain}) and understand their contextual usage, even when their visual appearances are highly similar. Moreover, this knowledge is inherently more balanced across categories, as it is learned from large-scale text corpora rather than the skewed distribution of a specific 3D dataset. As a result, \textbf{we claim that language priors provide a complementary and less biased source of 3D semantic supervision}, which is particularly beneficial for modeling tail classes that are underrepresented in visual data.

On top of these observations, we propose \textbf{\name{}}, a language-guided hierarchical learning framework for resolving long-tail ambiguity in unsupervised 3D point cloud segmentation. The key idea is to leverage the balanced world knowledge encoded in language models to counteract the bias of purely visual clustering.
Concretely, \name{} constructs an entity-level semantic prior using vision-language and segmentation models, and aggregates it into a global semantic bank. This prior is injected into the 3D feature space via contrastive alignment, enabling consistent associations between language-derived entities and 3D features.
Built upon this, we present a hierarchical clustering framework over superpoints with a dual-branch design that captures both local and global structures, producing multi-granularity pseudo-labels. Joint optimization with language-guided alignment encourages semantically consistent clustering and prevents tail classes from being absorbed by dominant categories.

We conduct extensive experiments on popular benchmarks, \ie, ScanNet-v2, S3DIS, and nuScenes, and \name{} significantly outperforms existing methods by +13.5, +12.9, and +8.9 mIoU, respectively. These results highlight the effectiveness of language priors in mitigating long-tail ambiguity and improving the segmentation of tail classes in 3D point clouds.

Our main contributions in this work are summarized as follows:
\begin{itemize}
\item We are \textit{the first} to leverage language priors to address the long-tail ambiguity problem in unsupervised 3D point cloud segmentation, providing a new perspective for mitigating clustering bias toward dominant classes.
\item We propose \name{}, a language-guided hierarchical learning framework that constructs an entity-level semantic prior from vision-language models and injects it into the clustering process via contrastive alignment, enabling balanced and semantically consistent representation learning.
\item We develop a dual-branch hierarchical clustering strategy that captures both local geometric consistency and global relational structure, generating multi-granularity pseudo-labels to improve clustering robustness.
\end{itemize}

\section{Related Work}
\noindent \textbf{Unsupervised 3D Semantic Segmentation.}
Directly adapting 2D unsupervised methods \cite{cho2021picie,ji2019invariant, caron2018deep} to 3D fails, primarily due to the inherent challenges of point sparsity and occlusion in 2D images. While self-supervised pretraining \cite{xie2020pointcontrast, hou2021exploring} captures discriminative geometry, it fundamentally focuses on low-level representation and lacks inherent high-level semantic understanding.
Recent unsupervised 3D segmentation methods \cite{chen2023pointdc, liu2024u3ds3, zhang2023growsp} bypass direct point-wise clustering by adopting superpoints or super-voxels guided iterative self-training. These approaches progressively refine pseudo-labels to distill low-level geometric coherence into high-level semantic discriminability. To mitigate the noise inherently associated with pseudo-label refinement, subsequent works introduce robust clustering mechanisms. LogoSP \cite{zhang2025logosp} constructs a global superpoint graph and leverages Graph Fourier Transform for frequency-domain clustering, while P-SLCR \cite{zhan2026p} employs a dynamic dual-prototype library to separate high-confidence points from uncertain ones, thereby filtering unreliable supervision signals. GrowSP++ \cite{zhang2026growsp++} uses a dual progressive growing strategy on superpoints and semantic primitives, evolving supervision from local geometry to global semantics to naturally filter early-stage errors.
However, these methods ignore minor class ambiguity. Class imbalance and sparsity, coupled with absent semantic priors, leave rare features vulnerable to collapse \cite{van2020scan, hamilton2022unsupervised}, as distance-based clustering inevitably absorbs them into dominant centroids. To address this, our hierarchical learning framework injects language priors into self-supervised learning, explicitly preserving long-tail representations during optimization.

\noindent \textbf{Cross-modal Priors Assisted 3D Understanding.}
Cross-modal priors extracted from Vision-Language Models (VLMs) \cite{radford2021learning} have emerged as a pivotal resource for advancing 3D understanding. By transferring the rich semantic knowledge encoded in image-text corpora to 3D representations \cite{xu2024controlrm,xu2026cyc3d,xu2021self,xu2021digging,chen2023costformer,xu2023semi,xu2024robustmvs}, these priors effectively assist 3D models in overcoming inherent data scarcity and weak semantic grounding. 
Recent works transfer cross-modal priors via multi-view projection \cite{zhu2023pointclip}, feature distillation \cite{chen2023pointdc, peng2023openscene}, and triplet alignment \cite{xue2023ulip, liu2023openshape}, enabling open-vocabulary and language-grounded 3D perception \cite{thengane2025foundational}. Yet, these paradigms overlook minor-class ambiguity in unsupervised clustering. Distance-based pseudo-labeling inherently marginalizes underrepresented categories, while open-vocabulary methods \cite{peng2023openscene, yang2024regionplc} rely on manual prompts or fine-tuning, leaving long-tail bias unresolved in purely label-free settings.
In contrast, our method constructs an offline entity-level semantic library by synergizing pre-extracted 3D entities with vision-language priors. By injecting these cross-modal priors into a contrastive learning pipeline, the learning of long-tail categories can be guaranteed.

\section{\name{}}
\begin{figure}[tbp]
  \centering
  \includegraphics[width=\linewidth]{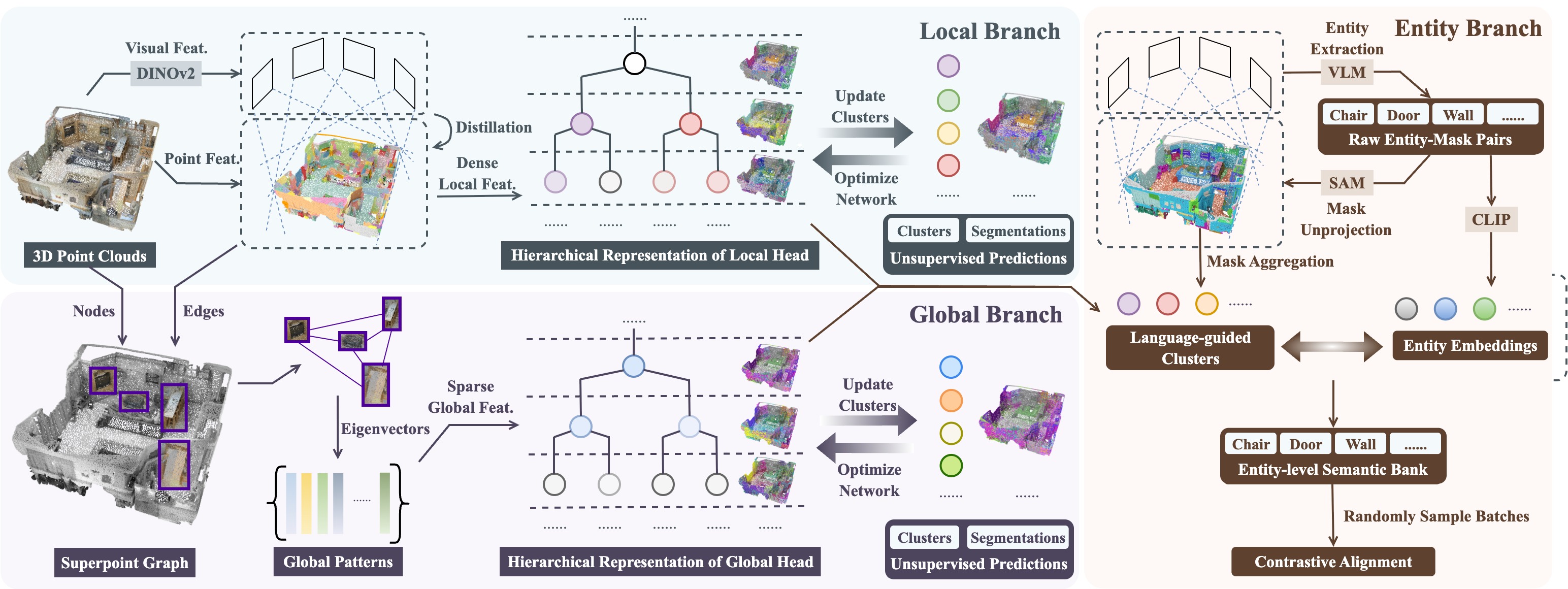}
  \vspace{-0.4cm}
  \caption{Overview of our \name{}. The Entity Branch inject balanced language priors into the 3D feature space via contrastive alignment. The Local and Global Branch respectively utilizes hierarchical learning-by-clustering to excavate the meaningful semantics from local regions and global patterns.}
  \vspace{-0.4cm}
  \label{fig:method}
\end{figure}

To handle long-tail ambiguity problem in unsupervised 3D point cloud segmentation, we propose \name{}, a language-guided hierarchical learning framework without requiring any manual annotations.
As shown in Fig. \ref{fig:method}, \name{} is comprised of three core branches: (1) Entity Branch (Sec. \ref{sec:entity}): It aims to utilize 2D foundation models (i.e. VLM \cite{guo2025seed1,seed2026seed1,yang2025qwen3}, SAM \cite{ravi2024sam} and clip \cite{radford2021learning}) to construct a multi-modal semantic bank in an offline manner, thereby injecting the balanced language priors with corresponding semantics into 3D features via contrastive learning. (2) Local Branch (Sec. \ref{sec:local}): It leverages the raw structure information from 3D point cloud to generate the inital superpoints, thus generating multi-granularity pseudo-labels hierarchically. The generated pseudo-labels are then fed back to optimize the network in an iterative manner. (3) Global Branch (Sec. \ref{sec:local}): It first groups the global representative patterns from 3D features and then conduct hierarchical learning-by-clustering iteration similar as Local Branch. Finally, the overview of the iterative learning-by-clustering pipeline is illustrated in Sec. \ref{sec:iter}.


\noindent\textbf{Problem Statement.}
Given a dataset with $M$ point clouds $\{\mP^1 \cdots \mP^i \cdots \mP^M\}$,  the $i$-th point cloud $\mP^i \in \mathbb{R}^{N_i \times 6}$ represents a scene point cloud with $N_i$ points of 6 dimensions including the XYZ coordinates and RGB intensities. For each point cloud $\mP^i$, there are $O$ RGB images $\{\mI^i_1 \cdots \mI^i_o \cdots \mI^i_O\}$ associated with the $i$-th point cloud. The 2D-3D correspondences between pixels and points can be easily built via the provided depth map and camera parameters in the 3D datasets. Our goal is to discover a set of semantically meaningful classes belonging to $c$ categories for any $n$-th point $\mP^i_n$ in the point cloud.

\noindent\textbf{Preliminary.}
We begin with the preliminaries of prior works \cite{zhang2023growsp,chen2023pointdc,zhang2025logosp,zhang2026growsp++} in a learning-by-clustering pipeline. 
The feature extractor of 3D point cloud is named as $f_\theta$ parameterized by $\theta$. The extracted feature is noted as $f_\theta (\mP_i) \in \mathbb{R}^{N \times D}$, where $D$ is the feature dimension. The baseline of learning-by-clustering in 3D point cloud segmentation can be summarized as follows:
\begin{enumerate}
    \item Optimizing the objective function and using K-Means to cluster the extracted features among all points in the dataset:
    \begin{equation}
        \small
        \min_{\bm{\mu},y} \sum_{i,j} \| f_\theta (\mP^i_j) - \bm{\mu} (y^i_j)\|_2^2 
        \label{eq1}
    \end{equation}
    where $\bm{\mu} \in \mathbb{R}^{c \times D}$ is the randomly initialized cluster centroids. $f_\theta (\mP^i_j)$ is the extracted feature of the $j$-th point in the $i$-th point cloud of the dataset. $y^i_j$ is the assigned label towards the cluster centroids of the $j$-th point in the $i$-th point cloud of the dataset.
    \item Use the clustered labels as pseudo-labels to train the backbone network $f_\theta$:
    \begin{equation}
        \small
        \min_{\theta} \sum_{i, j} \Ls_{CE} (g_{\bm{\mu}}(f_\theta(\mP^i_j)), y^i_j)
        \label{eq2}
    \end{equation}
    where $L_{CE}$ is the cross-entropy loss and $g_{\bm{\mu}}$ is a linear segmentation head parameterized by clustered centroids $\bm{\mu} \in \mathbb{R}^{c \times D}$.
\end{enumerate}

\subsection{Entity Branch}
\label{sec:entity}

To inject balanced semantic priors from language into 3D representation, we build an entity-level semantic bank using vision-language models and foundational segmentation models.
The entity-level semantic bank contains a huge amount of entity concepts in text and their corresponding 2D masks. The 3D masks are generated by unprojecting the 2D masks using the 2D-3D correspondence.
The aggregated 3D priors and entity-level semantics in the bank are then injected into the 3D feature space via contrastive alignment, bridging the associations between language-derived entities and 3D features.

\noindent\textbf{Raw Multi-Modal Priors.}
For a point cloud $\mP^i$ and its associated $K$ images, we feed each image to the VLM (Qwen3) and generate a list of entity categories presented in the image.
The entities are listed in the text response of VLM, such as "[Chair, Desk, ......]".
Then we separately input the entity-level text descriptions to the foundation segmentation model (SAM3) to generate the corresponding 2D masks.
Since one pixel might be covered by several different 2D masks, we filter out these labels with lower confidence from the foundation segmentation model, and retain only one single mask per pixel in the end.
The filtered 2D masks are then warped back to 3D space according to the 2D-3D correspondence between point clouds and images.
In this way, we can obtain paired samples comprised of entity-level text representations and the corresponding 3D masks.

\noindent\textbf{Entity-level Semantic Bank.} 
We aim to generate an entity-level semantic bank $\mB \in \mathbb{R}^{T \times C}$ from the extracted raw multi-modal priors in previous section.
$T$ is the total number of extracted entities and $C$ is the feature dimension of the 3D backbone $f_{\bm{\theta}}$.
Note that the feature dimension of the entity feature in the semantic bank should be the same as the 3D backbone rather than an arbitrary setting to conduct contrastive alignment in the following steps.
A naive thought is to build the semantic bank with pairwise entity-level text and 3D masks with CLIP embedding directly, but the raw entity-mask pairs might suffer from the vague descriptions (i.e., red bed vs blue bed, table vs desk).
These classes are annotated by the models as separate pairs but should not be treated as distinct categories during contrastive learning.
Consequently, we need to suppress the noisy and redundant feature representations in the entity-mask pairs before constructing the semantic bank.

Specifically, we first extract the CLIP embedding $\mF_e \in \mathbb{R}^{T \times 512}$ from the text descriptions of each entity (e.g., "a photo of a chair"). $T$ represents the total number of entities. The 3D mask of point cloud $\mP^i$ corresponding to the $t$-th entity text is noted as $\mM^{i,t}_e \in \mathbb{R}^{N_i}$. We can calculate the entity-level point feature $\mF_m \in \mathbb{R}^{T \times C}$ by:
\begin{equation}
\small
    \mF^t_m = \frac{1}{M} \sum_{i} (\frac{1}{| \mM^{i,t}_e |} \sum{f_{\bm{\theta}} (\mP^i) \cdot \mM^{i,t}_e})
    \label{eq3}
\end{equation}
where the extracted original feature $f_{\bm{\theta}} (\mP^i)$ from point cloud is averaged under the guidance of 3D entity mask $\mM^{i,t}_e$ to group the entity-level point features. $\mF^t_m$ represents the $t$-th of the $T$ entities in the entity-level point feature $\mF_m$.

To bridge the dimension gap between  $\mF_m \in \mathbb{R}^{T \times C}$ and $\mF_e \in \mathbb{R}^{T \times 512}$, we futher build a simple optimization problem on the entity-level point feature $\mF_m$:
\begin{equation}
\small
    \min_{\mF_m} \| \mG (\mF_m) - \mG (\mF_e)\|^2_2
    \label{eq4}
\end{equation}
where $\mG(\mX) = \mX \cdot \mX^T$ is the Gram matrix. We optimize the difference between gram matrices of entity-level point feature $\mF_m$ and entity-level text feature $\mF_e$ to inject the language prior to the point features.
This objective function can inherit the structure information of 3D features and merge the entity-level semantics guided by language into the 3D representation in an class agnostic manner.
After solving the optimization problem in Eq. \ref{eq4}, we can obtain the entity-level semantic bank $\mB = \mF_m^*$.

\noindent\textbf{Entity-level Contrastive Alignment.}
In consideration of the class agnosticism and the huge number of entities in semantic bank, we randomly sample a subset of entities from the semantic bank each time and conduct contrastive alignment between the sampled entities and the extracted feature from 3D backbone network $f_{\bm{\theta}}$.
Specifically, we use an InfoNCE-style contrastive loss to align the point features with the sampled entity prototypes in the bank during training:
\begin{equation}
\small
    \Ls_{\text{entity}} = - w_{c_i}  \frac{1}{M} \sum_i \log \frac{\exp \left(f_{\bm{\theta}} (\mP^i) \cdot \vb^{+} / \tau\right)}{\exp \left(f_{\bm{\theta}} (\mP^i) \cdot \vb^{+} / \tau\right) + \sum_{j=1}^{N_{\text{neg}}} \exp \left(f_{\bm{\theta}} (\mP^i) \cdot \vb_j^{-} / \tau\right)}
\end{equation}
where $\bm{b}^{+}$ is the prototype of the sampled batch, $\bm{b}^{-}_{j}$ is the hard negative samples (prototypes of other classes in the same batch).
$\tau$ is the temperature parameter.
$w_{c_i}$ is the balance weight for category ${c_i}$, calculated based on the inverse square root of the category frequency within the training batch: $w_{c_i} = 1 / \sqrt{n_{c_i}}$.
This loss guides the alignment of 3D features with the semantic space from pretraining, thereby achieving stronger semantic discrimination capabilities.

\subsection{Local Branch and Global Branch}
\label{sec:local}

\noindent\textbf{Cross-Modal Distillation Warmup.} Following previous works \cite{chen2023pointdc,zhang2025logosp}, we first extract pixel-level features with a pretrained visual foundation model (DINO-v2).
The pixel-level features are then mapped to the corresponding 3D points via the 2D-3D correspondence between images and point clouds.
As a warming up strategy, we distill the semantic information from these unprojected 2D features to 3D features with a cosine similarity loss.

\noindent\textbf{Hierarchical Representations.} To model the inherent hierarchical structure between different semantic concepts, we perform Ward-based hierarchical clustering directly on the grouped superpoints of 3D features.
Specifically, we can construct a hierarchical tree from the bottom to the top and truncate it at predefined level and number of clusters.
In default, we set the set of cluster numbers to $\gK = \{120, 80, K_{\text{prim}}\}$, where $K_{\text{prim}}$ is the number of target semantic primitives.


\noindent\textbf{Local Branch.} For each level of clusters $k \in \gK$, we initialize the weight matrix of a linear segmentation head with the centroids of clusters, parameterized by $\bm{\mu}^{\text{local}}_k \in \mathbb{R}^{k \times C}$. 
$k$ is the selected number of clusters of hierarchical tree and $C$ is the feature dimension.
The backbone network $f_{\bm{\theta}}$ is optimized via calculating the cross-entropy loss:
\begin{equation}
\small
    \Ls_{\text{local}} = \sum_{k \in \gK} \Ls_{CE} (g_{\bm{\mu}^{\text{local}}_k}(f_\theta(\mP^i_j)), \hat{y}^i_j)
\end{equation}
where $L_{CE}$ is the cross-entropy loss and $g_{\bm{\mu}^{\text{local}}_k}$ is the linear segmentation head parameterized by the cluster centroids $\bm{\mu}^{\text{local}}_k$. $\hat{y}^i_j$ is the assigned label towards the cluster centroids of the $j$-th point in the $i$-th point cloud of the dataset.

\noindent\textbf{Global Branch.} Although local branch can effectively group the adjacent regions with similar structure and semantics, they are prone to the limitations of local perspective and struggle to accurately group objects with the same semantics that are spatially separated. To address this issue, Global Branch incorporates spectral graph theory to capture the topological relationships and global contexts of 3D points. Following \cite{zhang2025logosp}, the global relationship among superpoints is constructed as a graph. After constructing and grouping global patterns, we obtain the refined graph fourier basis for each superpoint. More details are in the appendix~\ref{sec:graph_construct}.

Similar to Local Branch, we apply hierarchical clustering to generate the pseudo-labels first and then supervise the 3D backbone $f_\theta$ with cross-entropy loss:
\begin{equation}
\small
    \Ls_{\text{global}} = \sum_{k \in \gK} \Ls_{CE} (g_{\bm{\mu}^{\text{global}}_k}(f_\theta(\mP^i_j)), \tilde{y}^i_j)
\end{equation}
where $L_{CE}$ is the cross-entropy loss and $g_{\bm{\mu}^{\text{global}}_k}$ is the linear segmentation head parameterized by the cluster centroids. $\tilde{y}^i_j$ is the assigned label towards the cluster centroids of the $j$-th point in the $i$-th point cloud of the dataset.

\subsection{Iterative Optimization of \name{}}
\label{sec:iter}

The overall framework of \name{} is summarized as follows:

\begin{enumerate}
    \item Optimizing the objective function and using Ward-based hierarchical clustering to cluster the extracted features among all points in the dataset:
    \begin{equation}
    \small
        \min_{\bm{\mu}^{\text{local}}_k,y} \sum_{i,j} \| f_\theta (\mP^i_j) - \bm{\mu}^{\text{local}}_k (\hat{y}^i_j)\|_2^2
        \label{eq9}
    \end{equation}
    where $\bm{\mu}^{\text{local}}_k \in \mathbb{R}^{c \times D}$ is the randomly initialized cluster centroids of local features at $k \in \gK$ level. The clusters at different levels are aggregated under Ward linkage criterion. $f_\theta (\mP^i_j)$ is the extracted feature of the $j$-th point in the $i$-th point cloud of the dataset. $\hat{y}^i_j$ is the assigned label towards the cluster centroids of the $j$-th point in the $i$-th point cloud of the dataset.
    \begin{equation}
    \small
        \min_{\bm{\mu}^{\text{global}}_k,y} \sum_{i,j} \| f_\theta (\mP^i_j) - \bm{\mu}^{\text{global}}_k (\tilde{y}^i_j)\|_2^2 
        \label{eq10}
    \end{equation}
    where $\bm{\mu}^{\text{global}}_k \in \mathbb{R}^{c \times D}$ is the randomly initialized cluster centroids of global features at $k \in \gK$ level. The clusters at different levels are aggregated under Ward linkage criterion. $f_\theta (\mP^i_j)$ is the extracted feature of the $j$-th point in the $i$-th point cloud of the dataset. $\tilde{y}^i_j$ is the assigned label towards the cluster centroids of the $j$-th point in the $i$-th point cloud of the dataset.
    \item Use the clustered labels and semantic bank as pseudo-labels to train the backbone network $f_\theta$:
    \begin{equation}
    \small
        \min_{\theta} \sum_{i, j} \Ls_{\text{local}} + \Ls_{\text{global}} + \lambda \cdot \Ls_{\text{entity}}
        \label{eq11}
    \end{equation}
    $\lambda$ is a balancing hyperparameter. In each training iteration, gradients from all three losses ($\Ls_{\text{local}}$/$\Ls_{\text{global}}$/$\Ls_{\text{entity}}$) are backpropagated simultaneously to jointly optimize the parameters of the 3D backbone network.
\end{enumerate}

\section{Experiments}

To evaluate and compare with other methods, we choose two large-scale indoor datasets: (1) \textbf{ScanNet-v2} \cite{dai2017scannet} contains 1,613 scenes, among which 1,201 scenes are used as the training set, 312 scenes as the validation set, and the remaining 100 scenes as the online testing set. (2) \textbf{S3DIS} \cite{armeni2017joint} consists of six areas, divided into a total of 271 rooms. We select Area 5 as the validation set, with the other five areas serving as the training set; and a challenging autonomous driving dataset: (3) \textbf{nuScenes} \cite{caesar2020nuscenes}, as the outdoor dataset, contains 28,130 LiDAR scenes for training and 6018 scenes for validation. Following LogoSP, we choose mean Intersection-over-Union (mIoU), Overall Accuracy (OA), and mean Accuracy (mAcc) across all classes as our metrics.

\begin{figure*}[t]
    \centering
    \includegraphics[width=\textwidth, trim=0 40pt 0 0, clip]{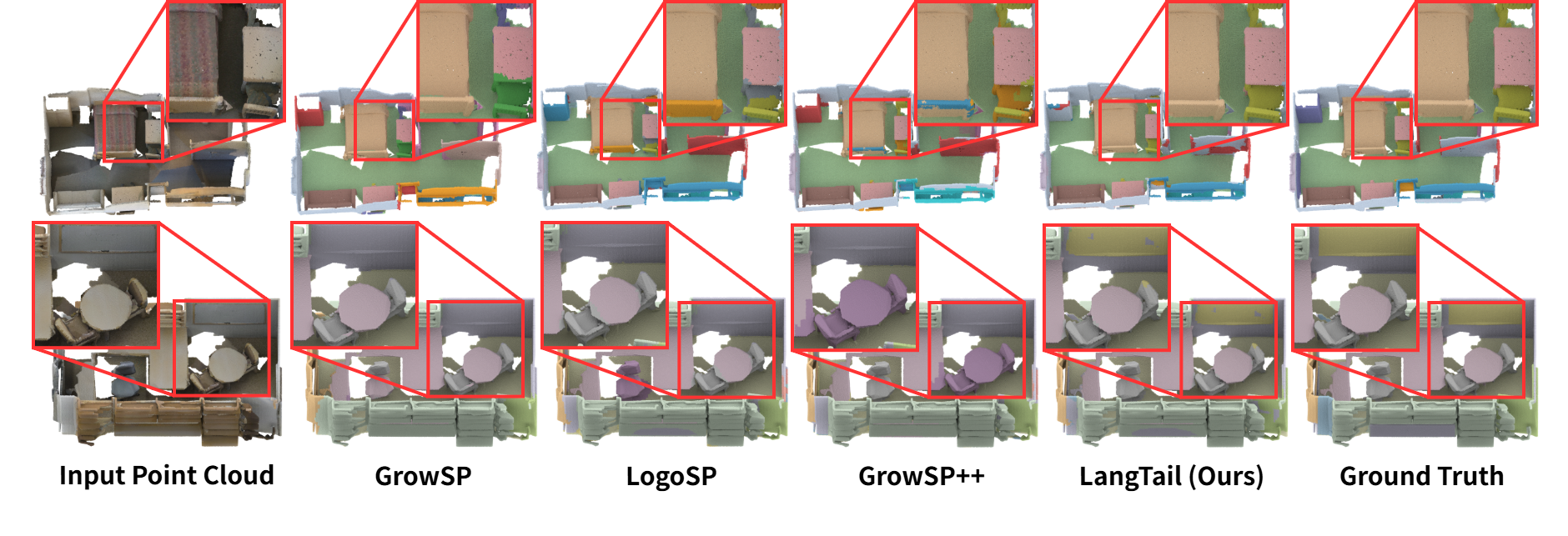}
    \vspace{-0.6cm}
    \caption{{Qualitative comparison on ScanNet (the first row) and S3DIS (the second row) datasets.}The red box highlights the differences}
    \label{fig:qualitative_results}
    \vspace{-0.3cm}
\end{figure*}

\begin{figure*}[t]
    \centering
    \includegraphics[width=\textwidth]{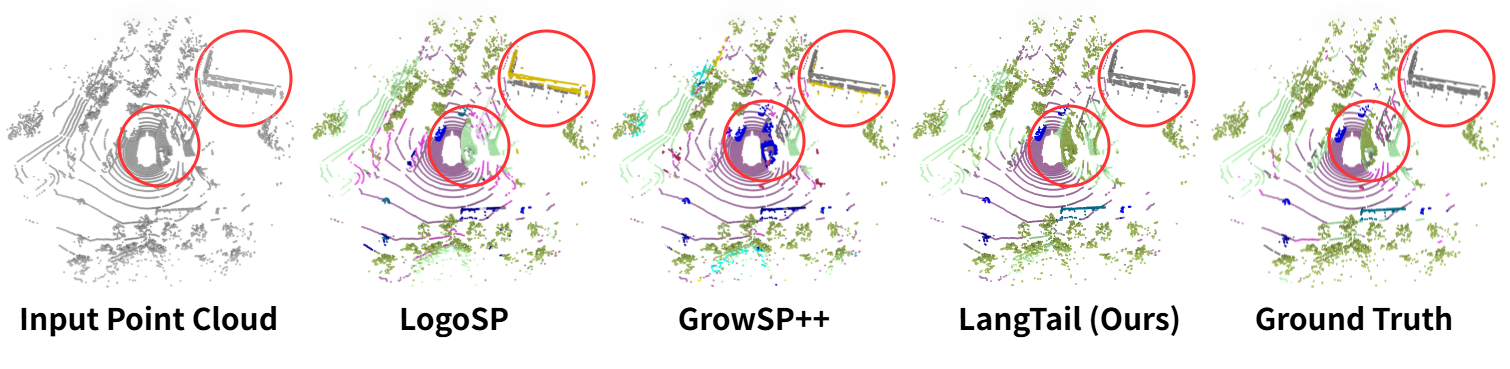}
    \vspace{-0.6cm}
    \caption{Qualitative comparison of point cloud segmentation on the nuScenes datasets.}
    \label{fig:nuscenes_qualitative_results}
    \vspace{-0.4cm}
\end{figure*}

\subsection{Evaluation on S3DIS}

\begin{wraptable}{r}{7.5cm}
\centering
\vspace{-0.7cm}
\caption{Results on ScanNet (Val) and S3DIS datasets.}
\label{tab:results_scannet_s3dis}
\resizebox{0.55\textwidth}{!}{
\begin{tabular}{r ccc ccc}
\toprule
\multirow{2}{*}{\textbf{Method}} & \multicolumn{3}{c}{\textbf{ScanNet (Val)}} & \multicolumn{3}{c}{\textbf{S3DIS}} \\
\cmidrule(lr){2-4} \cmidrule(lr){5-7}
& OA & mAcc & mIoU & OA & mAcc & mIoU \\
\midrule
\multicolumn{7}{l}{\textit{Supervised Methods}} \\
PointNet \cite{qi2017pointnet} & - & - & - & 77.5 & 59.1 & 44.6 \\
PointNet++ \cite{qi2017pointnet++} & - & - & - & 77.5 & 62.6 & 50.1 \\
SparseConv \cite{graham20183d} & - & - & - & 88.4 & 69.2 & 60.8 \\
\midrule
\multicolumn{7}{l}{\textit{Unsupervised Methods}} \\
IIC \cite{ji2019invariant} & 27.7 & 6.1 & 2.9 & 28.5 & 12.5 & 6.4 \\
IIC-S \cite{ji2019invariant} & 18.3 & 6.7 & 3.4 & 29.2 & 13.0 & 6.8 \\
IIC-PFH \cite{ji2019invariant} & 25.4 & 6.3 & 3.4 & 28.6 & 16.8 & 7.9 \\
IIC-S-PFH \cite{ji2019invariant} & 18.9 & 6.3 & 3.0 & 31.2 & 16.3 & 9.1 \\
PICIE \cite{cho2021picie} & 20.4 & 16.5 & 7.6 & 61.6 & 25.8 & 17.9 \\
PICIE-S \cite{cho2021picie} & 35.6 & 13.7 & 8.1 & 49.6 & 28.9 & 20.0 \\
PICIE-PFH \cite{cho2021picie} & 33.1 & 14.0 & 8.1 & 54.0 & 36.8 & 24.4 \\
PICIE-S-PFH \cite{cho2021picie} & 23.6 & 15.1 & 7.4 & 48.4 & 40.4 & 25.2 \\
PointDC \cite{chen2023pointdc} & 62.4 & 38.8 & 26.0 & 55.5 & 35.1 & 23.9 \\
PointDC-DINOv2 \cite{chen2023pointdc} & 64.7 & 45.0 & 29.6 & 75.7 & 48.7 & 40.2 \\
GrowSP \cite{zhang2023growsp} & 57.3 & 44.2 & 25.4 & 78.4 & 57.2 & 44.5 \\
LogoSP \cite{zhang2025logosp} & 64.7 & 50.8 & 35.8 & 82.8 & 55.9 & 46.5 \\
GrowSP++ \cite{zhang2026growsp++} & 70.5 & 48.8 & 33.2 & 78.7 & 60.1 & 46.6 \\
\textbf{LangTail (Ours)} & \textbf{78.7} & \textbf{59.2} & \textbf{46.7} & \textbf{86.7} & \textbf{69.6} & \textbf{59.5} \\
\bottomrule
\end{tabular}
}
\end{wraptable}

S3DIS \cite{armeni2017joint} comprises 6 areas and 271 rooms, with each point belonging to one of 12 categories. Following previous methods, we exclude clutter in loss computation or evaluation. We employed 6-fold cross-validation, training on 5 areas and evaluating on the remaining one.

\textbf{Results \& Analysis:}As shown in table ~\ref{tab:results_scannet_s3dis} , our method outperforms all baselines. Compared to the latest method GrowSP++ \cite{zhang2026growsp++}, our method achieves improvements of +8.0 in OA, +9.5 in mAcc, and +12.9 in mIoU. Furthermore, we outperform GrowSP++ on all categories, particularly in the board category, where accuracy improved from 0\% to 52.4\%. As shown in Figure~\ref{fig:qualitative_results}, our method not only yields cleaner boundaries but also accurately segments minor classes that baselines completely miss.

\subsection{Evaluation on ScanNet}
\begin{wraptable}{R}{5.0cm}
\centering
\vspace{-0.6cm}
\caption{Results on ScanNet (Test) dataset.}
\label{tab:results_scannet_test}
\begin{tabular}{l c}
\toprule
\textbf{Method} & \textbf{mIoU} \\
\midrule
\textit{Supervised Methods} & \\
PointNet++ \cite{qi2017pointnet++} & 33.9 \\
DGCNN \cite{wang2019dynamic} & 44.6 \\
PointCNN \cite{li2018pointcnn} & 45.8 \\
SparseConv \cite{graham20183d} & 72.5 \\
\midrule
\textit{Unsupervised Methods} & \\
GrowSP \cite{zhang2023growsp} & 26.9 \\
LogoSP \cite{zhang2025logosp} & 32.7 \\
GrowSP++ \cite{zhang2026growsp++} & 32.3 \\
\textbf{LangTail (Ours)} & \textbf{45.3} \\
\bottomrule
\end{tabular}
\vspace{-0.3cm}
\end{wraptable}
We conduct a comprehensive evaluation of our proposed method on ScanNet \cite{dai2017scannet}, a large-scale real-world dataset of indoor scenes. ScanNet consists of 1,201 scenes in the training set, 312 scenes in the validation set, and 100 scenes in the online test set. Each point in the dataset belongs to one of 20 categories or an undefined background. Following the approach of GrowSP++ \cite{zhang2026growsp++} and LogoSP \cite{zhang2025logosp}, we also employ the Hungarian matching algorithm to reorder the predicted labels for scoring, and report results separately for the offline validation set and the online hidden test set. The figure shows the qualitative results.

\textbf{Results \& Analysis:}As shown in table~\ref{tab:results_scannet_s3dis} and table~\ref{tab:results_scannet_test} ,we achieved state-of-the-art performance among current unsupervised methods. Compared to the latest method GrowSP++ \cite{zhang2026growsp++}, our method improves by +8.2 OA, +10.4 mAcc, and +13.5 mIoU. Our method not only achieves improvements on major categories but also performs exceptionally well on minor categories. For example, in the bathtub category, GrowSP++ achieves 0\% accuracy, whereas our method achieves 58.4\% accuracy. It demonstrates similarly outstanding performance on other minor categories such as sink, refrigerator, shower curtain etc., illustrating the effectiveness of our method in addressing the long-tail problem. Qualitative comparisons are visualized in Figure~\ref{fig:qualitative_results}.

\subsection{Evaluation on nuScenes}

\begin{wraptable}{r}{7.0cm}
\centering
\vspace{-0.7cm} 
\caption{Results on nuScenes dataset.}
\label{tab:results_nuscenes}
\resizebox{\linewidth}{!}{ 
\begin{tabular}{l ccc}
\toprule
\textbf{Method} & \textbf{OA} & \textbf{mAcc} & \textbf{mIoU} \\
\midrule
\textit{Unsupervised Methods} & & & \\
PointDC \cite{chen2023pointdc} & 56.8 & 29.4 & 17.7 \\
PointDC-DINOv2 \cite{chen2023pointdc} & 51.8 & 28.6 & 18.2 \\
GrowSP \cite{zhang2023growsp} & 39.2 & 17.5 & 10.2 \\
LogoSP \cite{zhang2025logosp} & 54.8 & 29.2 & 20.1 \\
GrowSP++ \cite{zhang2026growsp++} & 55.7 & 30.4 & 19.7 \\
\textbf{LangTail (Ours)} & \textbf{73.9} & \textbf{35.1} & \textbf{29.0} \\
\bottomrule
\end{tabular}
}
\vspace{-0.3cm}
\end{wraptable}

To demonstrate the generalizability of our method, we also conducted evaluation on nuScenes \cite{caesar2020nuscenes}, an outdoor dataset for autonomous driving. nuScenes is more challenging than the previous two datasets, comprising 28,130 LiDAR scenes for training and 6,018 scenes for validation. Each point belongs to one of 16 categories.

\textbf{Results \& Analysis:}As shown in table~\ref{tab:results_nuscenes}, our method outperforms all baselines and achieves state-of-the-art results. Compared to GrowSP++ \cite{zhang2026growsp++}, our method achieves +18.2 in OA, a +4.7 in mAcc, and +9.3 in mIoU, demonstrating the generalizability of our method in outdoor scenarios.The figure ~\ref{fig:nuscenes_qualitative_results} shows the qualitative result.

\subsection{Generalization to Unseen Datasets}

To verify whether the learned semantic features and multi-granularity prototypes possess generalization capabilities, we conducted a cross-dataset evaluation on two indoor datasets, ScanNet \cite{dai2017scannet} and S3DIS \cite{armeni2017joint}:
(1) \textbf{From ScanNet to S3DIS:} We directly use the model trained on ScanNet to extract features from ScanNet. Following a dual-branch hierarchical clustering process, we obtain a total of 440 cluster centroids, each with a granularity of $[120, 80, 20]$. These cluster centroids serve as classifiers for the S3DIS test regions.
(2) \textbf{From S3DIS to ScanNet:} Similarly, we directly use the model trained on S3DIS to extract features, yielding a total of 424 cluster centroids with branch-specific granularities of $[120, 80, 12]$. These cluster centroids serve as classifiers for the ScanNet test region.

    

\textbf{Results \& Analysis:}As shown in Table~\ref{tab:cross_dataset}, our model significantly outperforms all baseline methods in cross-dataset experiments. This validates that our method can also perform effectively on unfamiliar datasets.

\begin{table}[tbp]
\centering
\caption{Cross-dataset generalization evaluation. \textbf{Left (ScanNet \cite{dai2017scannet} $\rightarrow$ S3DIS \cite{armeni2017joint}):} Models are trained on the ScanNet dataset and tested on varying areas of S3DIS. \textbf{Right (S3DIS $\rightarrow$ ScanNet):} Models are trained on S3DIS and tested on ScanNet. For compactness, ``w/o A1'' denotes that Area 1 was excluded during training (i.e., trained on Areas 2-6). The best results are highlighted in \textbf{bold}.}
\label{tab:cross_dataset}
\resizebox{\textwidth}{!}{

\begin{tabular}{r ccccccc @{\hspace{4ex}} ccccccc} 
\toprule
\multirow{2}{*}{\textbf{Method}} & \multicolumn{7}{c}{\textbf{ScanNet $\rightarrow$ S3DIS (Tested on Area $i$)}} & \multicolumn{7}{c}{\textbf{S3DIS $\rightarrow$ ScanNet (Trained missing Area $i$)}} \\
\cmidrule(lr){2-8} \cmidrule(lr){9-15}
& A-1 & A-2 & A-3 & A-4 & A-5 & A-6 & \textbf{Mean} & w/o A1 & w/o A2 & w/o A3 & w/o A4 & w/o A5 & w/o A6 & \textbf{Mean} \\
\midrule
IIC \cite{ji2019invariant} & 3.7 & 3.8 & 3.8 & 4.0 & 3.8 & 3.7 & 3.8 & 3.5 & 3.4 & 3.7 & 3.5 & 3.5 & 3.6 & 3.5 \\
PiCIE \cite{cho2021picie} & 13.5 & 12.7 & 13.4 & 12.8 & 11.3 & 13.1 & 12.8 & 5.6 & 5.1 & 5.0 & 5.9 & 6.0 & 5.5 & 5.5 \\
GrowSP \cite{zhang2023growsp} & 24.2 & 21.9 & 26.1 & 25.0 & 23.7 & 27.9 & 24.8 & 16.9 & 17.8 & 16.4 & 16.1 & 17.1 & 15.3 & 16.6 \\
PointDC \cite{chen2023pointdc} & 23.6 & 20.9 & 24.6 & 19.5 & 20.1 & 29.7 & 23.1 & 10.3 & 10.1 & 10.4 & 8.4 & 10.0 & 9.9 & 9.9 \\
PointDC-DINOv2 \cite{chen2023pointdc} & 33.8 & 29.6 & 33.8 & 31.7 & 32.2 & 36.9 & 33.0 & 16.7 & 15.4 & 16.4 & 17.6 & 14.3 & 14.6 & 15.8 \\
LogoSP \cite{zhang2025logosp} & 43.8 & 37.5 & 47.0 & 40.7 & 44.9 & 47.9 & 43.6 & 16.9 & 16.9 & 16.8 & 16.8 & 16.7 & 17.0 & 16.9 \\
GrowSP++ \cite{zhang2026growsp++} & 39.5 & 32.9 & 41.7 & 36.0 & 36.1 & 43.8 & 38.3 & 17.5 & 16.4 & 17.2 & 16.1 & 17.5 & 16.6 & 16.9 \\
\textbf{LangTail (Ours)} & \textbf{48.3} & \textbf{42.6} & \textbf{50.1} & \textbf{44.3} & \textbf{47.7} & \textbf{49.7} & \textbf{47.1} & \textbf{19.4} & \textbf{19.0} & \textbf{19.6} & \textbf{18.6} & \textbf{18.6} & \textbf{19.7} & \textbf{19.2} \\
\bottomrule
\end{tabular}
}
\vspace{-0.3cm}
\end{table}

\begin{table}[tbp]
    \centering
    \caption{Comprehensive ablation study on the ScanNet validation set. We evaluate the impact of core modules (left) and varying hyperparameters (right). The complete setup is shown in the last row. }
    \label{tab:ablation}
    
    \resizebox{\linewidth}{!}{ 
    \begin{tabular}{c cccc cc c}
        \toprule
        \multirow{2}{*}{\textbf{Variant}} & \multicolumn{4}{c}{\textbf{Core Modules}} & \multicolumn{2}{c}{\textbf{Hyperparameters}} & \multirow{2}{*}{\textbf{mIoU (\%)}} \\
        \cmidrule(lr){2-5} \cmidrule(lr){6-7}
        & \textbf{Dual-Branch} & \textbf{Lang. Mod.} & \textbf{CLIP Align}  & \textbf{Hier. Clust.}& \textbf{Weight ($\lambda$)} & \textbf{Granularity ($K$)} & \\
        \midrule
        \multicolumn{8}{l}{\textit{Component Ablation (Progressive Addition)}} \\
        1 & \checkmark& & & & - & - & 36.0 \\
        2 & \checkmark &\checkmark & & & - & - & 36.4 \\
        3 & \checkmark & \checkmark & \checkmark& & - & - & 37.5 \\
        4 & \checkmark & \checkmark & \checkmark & \checkmark& - & [120, 80, 20] & 46.7 \\
        \midrule
        \multicolumn{8}{l}{\textit{Hyperparameter Ablation: Entity Loss Weight ($\lambda$)}} \\
        5 & \checkmark & \checkmark & \checkmark & \checkmark & 0.7 & [120, 80, 20] & 46.2 \\
        6 & \checkmark & \checkmark & \checkmark & \checkmark & 0.8 & [120, 80, 20] & 46.3 \\
        7 & \checkmark & \checkmark & \checkmark & \checkmark & 1.0 & [120, 80, 20] & 46.5 \\
        8 & \checkmark & \checkmark & \checkmark & \checkmark & 1.1 & [120, 80, 20] & 45.8 \\
        \midrule
        \multicolumn{8}{l}{\textit{Hyperparameter Ablation: Clustering Granularity ($K$)}} \\
        9 & \checkmark & \checkmark & \checkmark & \checkmark & 0.9 & [120, 40, 20] & 45.4 \\
        10 & \checkmark & \checkmark & \checkmark & \checkmark & 0.9 & [80, 40, 20] & 44.7 \\
        \midrule
        \textbf{Full} & \checkmark & \checkmark & \checkmark & \checkmark & \textbf{0.9} & \textbf{[120, 80, 20]} & \textbf{46.7} \\
        \bottomrule
    \end{tabular}
    }
    \vspace{-0.6cm}
\end{table}

\subsection{Ablation Study}
We conducted ablation experiments on the ScanNet validation set to evaluate the contribution of each component, with the following variants corresponding to the configurations in our table~\ref{tab:ablation} :

    
    
    
    
    
\textbf{(1)} To simultaneously capture semantic information across scales, we first integrate local and global branches. \textbf{(2)} Recognizing that purely visual supervision is limited in capturing open-world knowledge, we leverage Vision-Language Models (VLMs) and SAM to extract 2D masks.These masks are projected onto the 3D point cloud and processed by network $g$ to extract corresponding features.\textbf{(3)} Since the unsupervised nature of the task does not allow us to filter the generated vocabulary, the same object may have different entity names (e.g., “blue bed” and “bed”).To address this,we incorporate clip-based alignment to group objects of the same class in the feature space,providing reliable supervision signals.\textbf{(4)} To better mitigate the long-tail issue, we perform hierarchical clustering in both the local and global branches to extract prototypes at $\{120, 80, 20\}$ granularity levels, which are used to optimize the network.\textbf{(5)\textasciitilde(10)} $\lambda$ represents the weight of the entity loss, and $K$ denotes the clustering granularity.We performed ablation experiments on $\lambda \in \{0.7, 0.8, 0.9, 1.0, 1.1\}$ and on $K \in \{[120, 40, 20], [120, 80, 20], [80, 40, 20]\}$. In the full method,$\lambda$ and $K$ is consistently set to 0.9 and $[120, 80, 20]$,respectively.

\textbf{Results \& Analysis:}As shown in Table~\ref{tab:ablation}, \textbf{(1)} introducing a two-branch structure yields a score of 36.0 in mIoU, representing a slight improvement over LogoSP's 35.8. \textbf{(2)} Introducing the language module also improves the results, but the improvement is not significant due to a lack of alignment in the feature space. \textbf{(3)} After introducing CLIP alignment, the results show a more significant improvement, with marked progress in the segmentation of minor categories. \textbf{(4)} Finally, by introducing hierarchical clustering (Full), the mIoU score rose to 46.7, providing a better solution to the long-tail problem. \textbf{(5)} \textit{Robustness}: Our method is highly robust with regard to hyperparameter selection; the table shows the mIoU results under different hyperparameter settings. 

\section{Conclusion and Discussion}

LangTail incorporates real-world prior knowledge derived from Visual Language models, enabling it to recognize a broader range of categories. Following a dual-branch hierarchical clustering process, it captures multi-granularity pseudo-labels, thereby offering a novel solution to the long-tail ambiguity problem in unsupervised 3D point cloud segmentation.We conducted extensive experiments across multiple benchmarks, demonstrating the state-of-the-art performance of our method.Nevertheless, our method still has limitations in outdoor settings, which warrant further investigation.


{
\small
\bibliographystyle{IEEEtran}
\bibliography{reference}     
}






\newpage
\appendix
\section*{APPENDIX}
\section{Broader Imapacts}
\label{sec:broaderimapact}
Our work significantly advances unsupervised 3D semantic segmentation by enabling accurate, label-free discovery of complex scene structures directly from raw point clouds. It advances 3D scene understanding, offering clear positive societal impacts by eliminating the need for costly manual annotations and lowering the barrier for deploying 3D perception in robotics, autonomous navigation, augmented reality, and urban digital twins. Regarding negative impacts, we note that: (i) LangTail is presented as a research algorithm without direct deployment pathways or decision-making capabilities; (ii) the technology is modality-agnostic and does not process personal/sensitive data; (iii) its unsupervised nature actually reduces privacy risks by avoiding collection of user-specific labels. Any potential misuse would stem from application-level integration such as surveillance, not from the core algorithm itself.

\section{Implementation Details}
\label{sec:implementation}

\paragraph{Implementation Details.}
Consistent with previous methods, we employ a sparse convolutional neural network—Res16FPN18, implemented using MinkowskiEngine as the 3D backbone network. The input point cloud is converted into a voxel grid with a voxel size of 5 cm, and the final feature dimension of the network is set to 384. The network is initialized with distilled weights to accelerate convergence. For the training of the segmentation network, we use the AdamW optimizer with a batch size of 8 and train for a total of 200 epochs. The initial learning rate was set to $1 \times 10^{-4}$, and a Poly learning rate scheduling strategy (with an exponent of 0.9) was adopted to gradually reduce the learning rate to a minimum of $1 \times 10^{-8}$. To ensure computational efficiency, when the number of superpoints in the full dataset is excessive, we randomly sample up to 30,000 superpoints. All experiments are conducted on a single NVIDIA GeForce RTX 3090 GPU.

\paragraph{Evaluation Protocol.}
During the validation and testing phases, we concatenate the dynamically updated classifier weights from both branches across all granularities, resulting in dataset-specific number of pseudo-class prototypes (440 for ScanNet, 344 for S3DIS, and 352 for nuScenes) pseudo-class prototypes. The network assigns each point to one of these prototypes by calculating feature similarity. Subsequently, we use Hungarian matching to precisely map these pseudo-classes to the true semantic classes, and use this to compute the final evaluation metrics, including mean Intersection-over-Union (mIoU), mean Accuracy (mAcc), and Overall Accuracy (OA).

\section{Global superpoints graph construction and Frequency-domain Pattern Grouping}
\label{sec:graph_construct}
\paragraph{Global Superpoint Graph Construction}
 Let $\mF \in \sR^{S \times C}$ (where $S$ is the number of superpoints $C=384$) denote the initial feature matrix of all superpoints.We construct a global graph $\gG$ where each node represents a superpoint. The edge weight $a_{ij}$ between the $i$-th and $j$-th nodes is defined as follows:
\begin{equation}
    a_{ij} = \exp(-\|\vf_i - \vf_j\|^2),
\end{equation}
where $\vf_i, \vf_j$ are the corresponding feature vectors from $\mF$. We then compute the normalized Laplacian matrix $\mL = \mD^{-1/2}(\mD - \mA)\mD^{-1/2}$, where $\mA$ is the adjacency matrix and $\mD$ is the degree matrix. By performing eigen-decomposition on $\mL$:
\begin{equation}
    \mL = \mU \mLambda \mU^\top, \quad \mU = [\vu_1, \dots, \vu_s, \dots, \vu_S],
\end{equation}
where the eigenvectors $\mU \in \sR^{S \times S}$ serve as the graph Fourier basis. Each eigenvector $\vu_s \in \sR^{S \times 1}$ represents a specific global structural pattern defined over all $S$ superpoints.

\paragraph{Frequency-domain Pattern Grouping}
Directly using all $S$ global patterns is computationally expensive and contains semantic redundancies. To obtain discriminative semantic priors, we group these patterns as follows:

\begin{itemize}
    \item \textbf{Spectral Projection:} We project the spatial features $\mF$ into the graph frequency domain to obtain frequency features $\mF_{feq} \in \sR^{S \times C}$:
    \begin{equation}
        \mF_{feq} = \mU^\top \mF, \quad \mF_{feq} = [\vf_{feq, 1}, \dots, \vf_{feq, s}, \dots, \vf_{feq, S}]^\top,
    \end{equation}
    where each row $\vf_{feq, s} \in \sR^{1 \times C}$ represents the frequency response corresponding to the $s$-th global pattern $\vu_s$.
    
    \item \textbf{K-means Clustering:} We apply K-means clustering to the frequency vectors $\{\vf_{feq, 1}, \dots, \vf_{feq, S}\}$ to partition the $S$ original patterns into $S'$ clusters.
    
    \item \textbf{Pattern Aggregation:} For each cluster, we average its assigned eigenvectors to form a refined global pattern $\vv_{s'}$:
    \begin{equation}
        \vv_{s'} = \frac{1}{W} \sum_{w=1}^{W} \vu_w, \quad \vv_{s'} \in \sR^{S \times 1},
    \end{equation}
    where $W$ is the number of original patterns in the $s'$-th cluster. 
\end{itemize}

The resulting matrix $\mV = [\vv_1, \dots, \vv_{S'}] \in \sR^{S \times S'}$ captures consolidated semantic priors by grouping patterns with similar spectral responses. This refined basis $\mV$ is subsequently used for global branch.

\section{Additional Quantitative Results}
\label{sec:quantitative results}
\paragraph{ScanNet.}
Tables ~\ref{tab:scannet_results_val} and Table ~\ref{tab:scannet_results_test} provide detailed results for each category on the validation and hidden test sets. Our method significantly outperforms all unsupervised baseline models, particularly in challenging subcategories such as \textit{sinks}, \textit{bathtubs}, and \textit{refrigerators}.

\begin{table}[h]
\centering
\caption{Quantitative results on the validation split of ScanNet dataset. All 20 categories are evaluated.}
\label{tab:scannet_results_val}
\resizebox{\textwidth}{!}{
\begin{tabular}{l ccccccccccccccccccccccc}
\toprule
Method & OA(\%) & mAcc(\%) & mIoU(\%) & wall. & floor. & cab. & bed. & chair. & sofa. & table & door. & wind. & books. & pic. & counter. & desk. & curtain. & fridge. & shower. & toilet. & sink. & bathtub. & otherf. \\
\midrule
RandCNN          & 11.9   & 8.4      & 3.2      & 9.3   & 10.0   & 3.5  & 2.5  & 6.8    & 2.0   & 4.8   & 5.1   & 3.9   & 3.1    & 1.7  & 0.8      & 2.3   & 2.8      & 1.0     & 0.2     & 0.2     & 0.1   & 0.6      & 3.6     \\
van Kmeans       & 10.1   & 10.0     & 3.4      & 9.0   & 9.8    & 3.2  & 2.9  & 5.5    & 3.3   & 4.3   & 3.5   & 5.5   & 3.3    & \textbf{2.6}  & 0.8      & 2.9   & 4.3      & 0.8     & 0.7     & 0.8     & 0.3   & 0.9      & 4.0     \\
van Kmeans-S     & 10.2   & 9.8      & 3.4      & 8.9   & 10.3   & 3.4  & 3.2  & 5.5    & 3.4   & 4.2   & 3.4   & 5.2   & 3.1    & \textbf{2.6}  & 0.7      & 2.8   & 4.2      & 0.6     & 0.8     & 0.7     & 0.2   & 1.0      & 4.1     \\
van Kmeans-PFH   & 10.4   & 10.3     & 3.5      & 8.6   & 12.7   & 2.9  & 2.8  & 4.5    & 3.2   & 3.6   & 3.7   & 6.3   & 4.0    & 2.4  & 1.0      & 2.9   & 3.2      & 1.0     & 1.0     & 0.6     & 0.4   & 1.1      & 3.5     \\
van Kmeans-S-PFH & 12.2   & 9.3      & 3.6      & 11.3  & 12.3   & 2.9  & 2.4  & 5.4    & 2.8   & 4.2   & 3.8   & 5.8   & 3.8    & 2.3  & 1.2      & 2.4   & 2.9      & 0.9     & 1.4     & 0.6     & 0.1   & 1.1      & 4.1     \\
IIC \cite{ji2019invariant}              & 27.7   & 6.1      & 2.9      & 25.3  & 20.5   & 0.6  & 0.3  & 3.7    & 0.4   & 1.3   & 1.3   & 1.1   & 1.9    & 0.2  & 0.1      & 0.6   & 0.3      & 0.4     & 0       & 0       & 0     & 0.2      & 0.5     \\
IIC-S \cite{ji2019invariant}            & 18.3   & 6.7      & 3.4      & 18.3  & 16.0   & 2.6  & 2.3  & 4.4    & 2.0   & 5.4   & 3.2   & 2.9   & 3.3    & 0.7  & 0.4      & 1.4   & 1.6      & 0.7     & 0.1     & 0.3     & 0.1   & 0        & 2.6     \\
IIC-PFH \cite{ji2019invariant}          & 25.4   & 6.3      & 3.4      & 29.6  & 14.9   & 1.1  & 1.0  & 5.6    & 0.8   & 3.6   & 3.0   & 1.6   & 1.3    & 0    & 0.3      & 1.0   & 0.4      & 0.4     & 0.2     & 0       & 0.1   & 0        & 3.2     \\
IIC-S-PFH \cite{ji2019invariant}        & 18.9   & 6.3      & 3.0      & 18.0  & 15.9   & 3.4  & 0.9  & 7.1    & 0.6   & 0.8   & 4.3   & 1.6   & 3.5    & 0.4  & 0.1      & 0.3   & 0.3      & 0       & 0       & 0.1     & 0     & 0        & 2.7     \\
PICIE \cite{cho2021picie}            & 20.4   & 16.5     & 7.6      & 14.7  & 24.5   & 6.3  & 5.2  & 18.0   & 8.4   & 33.2  & 6.7   & 4.8   & 9.3    & 2.1  & 0.1      & 2.7   & 8.0      & 1.1     & 2.1     & 0       & 0     & 0.5      & 5.0     \\
PICIE-S \cite{cho2021picie}          & 35.6   & 13.7     & 8.1      & 38.4  & 53.9   & 4.3  & 2.7  & 10.2   & 6.3   & 14.1  & 5.2   & 4.0   & 6.0    & 0.2  & 1.3      & 2.1   & 1.5      & 0.2     & 0       & 2.6     & 3.1   & 1.3      & 4.3     \\
PICIE-PFH \cite{cho2021picie}        & 33.1   & 14.0     & 8.1      & 34.7  & 54.8   & 3.9  & 5.4  & 13.3   & 6.5   & 11.7  & 4.2   & 3.8   & 6.5    & 0.5  & 1.0      & 2.6   & 5.0      & 1.3     & 1.0     & 0.6     & 0     & 0.7      & 4.4     \\
PICIE-S-PFH \cite{cho2021picie}      & 23.6   & 15.1     & 7.4      & 18.1  & 39.1   & 5.4  & 4.9  & 13.4   & 6.9   & 20.3  & 5.8   & 4.5   & 7.7    & 1.2  & 3.0      & 5.8   & 4.7      & 0.6     & 1.2     & 0.4     & 0     & 1.1      & 4.5     \\
PointDC \cite{chen2023pointdc}          & 62.4   & 38.8     & 26.0     & 59.1  & \textbf{94.0} & 22.0 & 43.2 & 30.4   & 35.9  & 38.3  & 14.3  & 37.4  & 44.4   & 1.2  & 2.4      & 2.3   & 39.4     & 2.0     & 0       & 38.6    & 0     & 2.2      & 12.7    \\
PointDC-DINOv2 \cite{chen2023pointdc}   & 64.7   & 45.0     & 29.6     & 57.0  & 86.0   & 18.5 & 60.6 & 60.1   & 46.2  & 46.4  & 27.0  & 39.0  & 54.2   & 0.7  & 25.0     & 18.1  & 22.7     & 0.2     & 2.8     & 16.8    & 0     & 0        & 10.4    \\
GrowSP \cite{zhang2023growsp}           & 57.3   & 44.2     & 25.4     & 40.7  & 89.8   & 24.0 & 47.2 & 45.5   & 43.0  & 39.4  & 14.1  & 20.0  & 53.5   & 0.1  & 5.4      & 13.3  & 8.4      & 2.1     & 11.3    & 20.6    & 19.4 & 0   & 9.8     \\
LogoSP \cite{zhang2025logosp}           & 64.7   & 50.8     & 35.8     & 46.3  & 86.6   & 20.7 & 66.8 & 63.3   & 50.9 & 47.1  & 33.8  & 41.6  &62.8   & 1.0  & 38.0     & 10.5  & 28.6     & 0.5     & 0.0     & 46.3    & 0.0   & 42.3     & 29.6    \\
GrowSP++ \cite{zhang2026growsp++}         & 70.5   & 48.8     & 33.2     & 63.6  & 87.9   & 21.9 & 65.3 & 63.6   & 49.8  & 45.0  & 15.5  & 42.9  & 61.8   & 0.1  & 35.7     & 27.4  & 34.9     & 2.1     & 7.8     & 26.3    & 0     & 0        & 12.0    \\
LangTail(Ours)             & \textbf{78.7} & \textbf{59.2} & \textbf{46.7} & \textbf{69.7} & 90.0 & \textbf{42.1} & \textbf{67.9} & \textbf{69.8} & \textbf{52.8}  & \textbf{51.1} & \textbf{37.9}  & \textbf{44.2}  & \textbf{64.1}   & 0.0  & \textbf{43.5} & \textbf{40.2} & \textbf{40.0} & \textbf{3.5} & \textbf{40.1} & \textbf{67.7} & \textbf{20.9}  & \textbf{52.4} & \textbf{35.5}   \\
\bottomrule
\end{tabular}
}
\end{table}

\begin{table*}[h]
\centering
\caption{Quantitative results on the online hidden split of ScanNet dataset. All 20 categories are evaluated.}
\label{tab:scannet_results_test}
\resizebox{\textwidth}{!}{ 
\begin{tabular}{ll ccccccccccccccccccccc}
\toprule
\multicolumn{2}{l}{Method} & mIoU(\%) & wall. & floor. & cab. & bed. & chair. & sofa. & table & door. & wind. & books. & pic. & counter. & desk. & curtain. & fridge. & shower. & toilet. & sink. & bathtub. & otherf. \\ 
\midrule
\multirow{4}{*}{\begin{tabular}[c]{@{}l@{}}Supervised \\ Methods\end{tabular}} 
& PointNet++ \cite{qi2017pointnet++} & 33.9 & 33.9 & 52.3 & 67.7 & 25.6 & 47.8 & 36.0 & 34.6 & 23.2 & 26.1 & 25.2 & 45.8 & 11.7 & 25.0 & 27.8 & 24.7 & 18.3 & 14.5 & 54.8 & 36.4 & 58.4 \\
& DGCNN \cite{wang2019dynamic}      & 44.6 & 44.6 & 72.3 & 93.7 & 36.6 & 62.3 & 65.1 & 57.7 & 44.5 & 33.0 & 39.4 & 46.3 & 12.6 & 31.0 & 34.9 & 38.9 & 28.5 & 22.4 & 62.5 & 35.0 & 47.4 \\
& PointCNN \cite{li2018pointcnn}   & 45.8 & 45.8 & 70.9 & 94.4 & 32.1 & 61.1 & 71.5 & 54.5 & 45.6 & 31.9 & 47.5 & 35.6 & 16.4 & 29.9 & 32.8 & 37.6 & 21.6 & 22.9 & 75.5 & 48.4 & 57.7 \\
& SparseConv \cite{graham20183d} & 72.5 & 72.5 & 86.5 & 95.5 & 72.1 & 82.1 & 86.9 & 82.3 & 62.8 & 61.4 & 68.3 & 84.6 & 32.5 & 53.3 & 60.3 & 75.4 & 71.0 & 87.0 & 93.4 & 72.4 & 64.7 \\
\midrule
\multirow{4}{*}{\begin{tabular}[c]{@{}l@{}}Unsupervised \\ Methods\end{tabular}} 
& GrowSP     & 26.9 & 26.9 & 32.8 & \textbf{89.6} & 15.2 & 62.9 & 55.3 & 38.9 & 32.0 & 14.4 & 23.0 & \textbf{59.9} & 0.0 & 12.5 & 11.4 & 6.1 & 1.2 & 9.3 & \textbf{43.9} & 14.0 & 0.0 \\
& LogoSP \cite{zhang2025logosp}     & 32.7 & 41.4 & 87.1 & 18.1 & \textbf{68.4} & 56.2 & 49.9 & 39.6 & 30.2 & \textbf{48.7} & \textbf{49.2} & 0.1 & \textbf{29.1} & 7.3 & 33.4 & 0.0 & 0.0 & 54.3 & 0.0 & 21.1 & 19.3 \\
& GrowSP++ \cite{zhang2026growsp++}   & 32.3 & 32.3 & 66.4 & 86.5 & 14.7 & 58.9 & \textbf{55.5} & \textbf{46.2} & \textbf{38.1} & 26.2 & 47.3 & 49.9 & 0.0 & 29.0 & 29.0 & 33.6 & 0.0 & 0.0 & 38.9 & 0.0 & 11.4 \\
& ours       & \textbf{45.3} & \textbf{70.8} & \textbf{89.5} & 36.0 & 63.1 & \textbf{67.1} & 55.2 & 39.8 & 34.7 & 46.9 & 46.8 & 0.0 & 28.6 & \textbf{36.9} & \textbf{47.9} & \textbf{33.8} & \textbf{47.9} & \textbf{68.2} & 16.7 & \textbf{48.0} & \textbf{27.1} \\
\bottomrule
\end{tabular}
}
\end{table*}

\paragraph{S3DIS.}
 The per-category results for each area evaluation are presented in Table~\ref{tab:S3DIS_area1} to Table~\ref{tab:S3DIS_area6} .Our method outperformed the unsupervised baseline method in all experiments.It has also achieved greatest performance in many specific categories
\begin{table*}[h]
\centering
\caption{Quantitative results of our method and baselines on the Area-1 of S3DIS dataset.}
\label{tab:S3DIS_area1}
\resizebox{\textwidth}{!}{ 
\begin{tabular}{ll ccccccccccccccc}
\toprule
\multicolumn{2}{l}{Method} & OA(\%) & mAcc(\%) & mIoU(\%) & ceil. & floor & wall & beam & col. & wind. & door & table & chair & sofa & book. & board. \\ 
\midrule
\multirow{3}{*}{\begin{tabular}[c]{@{}l@{}}Supervised \\ Methods\end{tabular}} 
& PointNet \cite{qi2017pointnet}      & 75.4 & 74.8 & 55.0 & 88.3 & 93.2 & 69.2 & 49.5 & 37.8 & 74.5 & 65.6 & 41.2 & 42.5 & 22.3 & 35.4 & 40.9 \\
& PointNet++ \cite{qi2017pointnet++}    & 76.1 & 77.9 & 58.2 & 90.5 & 94.4 & 65.7 & 38.2 & 31.9 & 61.5 & 66.0 & 45.3 & 60.4 & 41.2 & 45.8 & 57.4 \\
& SparseConv \cite{graham20183d}    & 89.0 & 79.5 & 72.5 & 93.6 & 95.6 & 76.1 & 65.9 & 60.9 & 60.0 & 74.2 & 81.9 & 85.4 & 69.2 & 73.4 & 33.5 \\
\midrule
\multirow{19}{*}{\begin{tabular}[c]{@{}l@{}}Unsupervised \\ Methods\end{tabular}} 
& RandCNN          & 24.9 & 20.3 & 10.9 & 27.8 & 29.0 & 16.6 & 10.7 & 4.7 & 14.8 & 8.1 & 4.1 & 3.7 & 0.2 & 6.4 & 5.1 \\
& van Kmeans       & 20.9 & 24.1 & 10.1 & 15.4 & 17.8 & 10.5 & 16.8 & 1.9 & 16.0 & 12.1 & 9.9 & 8.1 & 0.1 & 6.2 & 6.7 \\
& van Kmeans-S     & 21.8 & 25.6 & 10.8 & 18.1 & 20.4 & 9.5 & 14.5 & 1.0 & 17.2 & 16.2 & 9.6 & 6.8 & 0.6 & 8.2 & 7.6 \\
& van Kmeans-PFH   & 26.4 & 25.7 & 12.6 & 34.0 & 27.4 & 11.6 & 17.0 & 5.2 & 15.4 & 9.7 & 10.8 & 5.8 & 1.6 & 8.1 & 4.3 \\
& van Kmeans-S-PFH & 25.1 & 23.0 & 10.8 & 26.1 & 18.5 & 15.0 & 4.6 & 2.6 & 20.5 & 15.9 & 10.4 & 6.0 & 0.6 & 5.9 & 3.7 \\
& IIC \cite{ji2019invariant}              & 29.2 & 14.3 & 8.0 & 17.0 & 31.4 & 25.6 & 4.3 & 11.1 & 0.0 & 2.6 & 1.4 & 0.7 & 0.0 & 0.2 & 1.4 \\
& IIC-S \cite{ji2019invariant}            & 31.3 & 21.1 & 8.2 & 35.5 & 11.6 & 13.9 & 1.3 & 0.2 & 13.0 & 15.6 & 5.1 & 0.6 & 0.3 & 0.3 & 7.6 \\
& IIC-PFH \cite{ji2019invariant}          & 27.3 & 12.0 & 6.1 & 17.1 & 17.0 & 23.4 & 4.4 & 0.3 & 0.8 & 4.0 & 0.4 & 0.4 & 0.0 & 4.9 & 0.1 \\
& IIC-S-PFH \cite{ji2019invariant}        & 24.1 & 13.2 & 7.2 & 21.7 & 16.8 & 18.0 & 5.0 & 3.6 & 2.9 & 3.7 & 4.4 & 0.7 & 0.0 & 4.7 & 4.5 \\
& PICIE \cite{cho2021picie}            & 45.7 & 28.3 & 19.4 & 77.2 & 63.1 & 24.5 & 15.8 & 3.3 & 4.4 & 9.6 & 10.2 & 14.7 & 0.0 & 9.9 & 0.0 \\
& PICIE-S \cite{cho2021picie}          & 48.9 & 30.3 & 21.8 & 77.7 & 58.1 & 39.0 & 13.6 & 2.5 & 2.8 & 11.8 & 20.0 & 17.1 & 1.8 & 17.1 & 0.6 \\
& PICIE-PFH \cite{cho2021picie}        & 51.9 & 38.4 & 27.1 & 59.8 & 78.7 & 37.5 & 6.8 & 9.5 & 4.9 & 26.4 & 50.2 & 29.5 & 0.0 & 22.8 & 0.1 \\
& PICIE-S-PFH \cite{cho2021picie}      & 47.9 & 38.9 & 27.2 & 71.5 & 66.3 & 23.9 & 20.4 & 8.8 & 6.9 & 12.9 & 39.3 & 32.2 & 0.2 & 42.0 & 2.4 \\
& PointDC \cite{chen2023pointdc}          & 58.0 & 41.5 & 28.8 & 88.7 & 89.5 & 31.9 & 1.5 & 7.1 & 17.6 & 12.4 & 46.4 & 17.0 & 0.0 & 32.7 & 0.0 \\
& PointDC-DINOv2 \cite{chen2023pointdc}   & 73.8 & 55.4 & 44.0 & 85.7 & \textbf{93.7} & 58.9 & 10.1 & \textbf{20.2} & 0.0 & 45.0 & 70.1 & 61.5 & 44.0 & 39.4 & 0.0 \\
& GrowSP \cite{zhang2023growsp}           & 72.9 & 60.4 & 45.6 & \textbf{94.2} & 90.8 & 52.7 & 36.7 & 19.7 & 33.3 & 35.8 & 66.5 & 72.6 & 13.1 & 31.2 & 16.7 \\
& LogoSP \cite{zhang2025logosp}           & 76.9 & 60.0 & 48.9 & 89.0 & 93.2 & 63.2 & 27.5 & 19.2 & 71.3 & 39.7 & 69.7 & 69.2 & 0.7 & 43.6 & 0.0 \\
& GrowSP++ \cite{zhang2026growsp++}         & 72.9 & 62.9 & 47.5 & 78.8 & 92.9 & 53.2 & 0.0 & 1.1 & 74.4 & 33.9 & 66.8 & 71.4 & 54.8 & 43.1 & 0.1 \\
& \textbf{LangTail(ours)}              & \textbf{84.3} & \textbf{74.6} & \textbf{63.6} & 92.8 & 90.9 & \textbf{68.7} & \textbf{45.0} & 1.8 & \textbf{76.3} & \textbf{66.1} & \textbf{76.3} & \textbf{79.4} & \textbf{67.5} & \textbf{62.5} & \textbf{35.6} \\
\bottomrule
\end{tabular}
}
\end{table*}

\begin{table*}[h]
\centering
\caption{Quantitative results of our method and baselines on the Area-2 of S3DIS dataset.}
\label{tab:S3DIS_area2}
\resizebox{\textwidth}{!}{
\begin{tabular}{ll ccccccccccccccc}
\toprule
\multicolumn{2}{l}{Method} & OA(\%) & mAcc(\%) & mIoU(\%) & ceil. & floor & wall & beam & col. & wind. & door & table & chair & sofa & book. & board. \\ 
\midrule
\multirow{3}{*}{\begin{tabular}[c]{@{}l@{}}Supervised \\ Methods\end{tabular}} 
& PointNet \cite{qi2017pointnet}      & 72.5 & 55.5 & 36.6 & 79.2 & 87.4 & 64.9 & 14.5 & 8.2 & 14.8 & 39.6 & 28.8 & 64.0 & 7.8 & 24.4 & 5.1 \\
& PointNet++ \cite{qi2017pointnet++}    & 72.1 & 62.3 & 39.9 & 85.8 & 69.6 & 71.2 & 24.9 & 27.5 & 32.5 & 43.6 & 27.4 & 51.3 & 6.0 & 26.8 & 12.4 \\
& SparseConv \cite{graham20183d}    & 87.9 & 69.5 & 57.3 & 89.5 & 93.8 & 77.0 & 29.1 & 32.5 & 65.5 & 45.7 & 67.9 & 88.8 & 34.9 & 54.5 & 8.15 \\
\midrule
\multirow{19}{*}{\begin{tabular}[c]{@{}l@{}}Unsupervised \\ Methods\end{tabular}} 
& RandCNN          & 20.4 & 17.0 & 7.6 & 16.8 & 19.2 & 12.8 & 1.6 & 1.1 & 1.5 & 9.4 & 2.9 & 16.1 & 0.3 & 6.4 & \textbf{13.0} \\
& van Kmeans       & 17.6 & 16.6 & 6.4 & 16.4 & 15.6 & 11.3 & 3.3 & 0.9 & 0.4 & 6.8 & 3.7 & 11.0 & 1.4 & 4.6 & 1.5 \\
& van Kmeans-S     & 17.3 & 17.6 & 6.3 & 14.7 & 15.0 & 12.1 & 3.0 & 0.9 & 0.0 & 7.0 & 3.6 & 11.4 & 1.9 & 4.3 & 2.0 \\
& van Kmeans-PFH   & 21.8 & 22.3 & 9.0 & 31.0 & 21.1 & 10.1 & 3.0 & 2.6 & 3.2 & 10.7 & 5.3 & 9.8 & 4.0 & 7.2 & 0.6 \\
& van Kmeans-S-PFH & 21.5 & 18.0 & 8.0 & 23.7 & 21.1 & 14.1 & 3.8 & 1.8 & 1.8 & 5.4 & 4.3 & 10.2 & 2.5 & 6.5 & 0.9 \\
& IIC \cite{ji2019invariant}              & 41.6 & 16.8 & 10.6 & 33.0 & 43.7 & 27.6 & 1.7 & 0.0 & 0.0 & 5.6 & 0.1 & 13.0 & 0.0 & 2.8 & 0.0 \\
& IIC-S \cite{ji2019invariant}            & 30.3 & 11.9 & 6.2 & 29.3 & 7.6 & 22.6 & 0.5 & 1.5 & 0.0 & 3.5 & 0.3 & 2.2 & 1.8 & 3.4 & 1.1 \\
& IIC-PFH \cite{ji2019invariant}          & 27.3 & 10.5 & 5.8 & 16.0 & 13.9 & 24.9 & 0.3 & 0.1 & 0.0 & 4.0 & 0.6 & 8.3 & 0.1 & 0.8 & 0.1 \\
& IIC-S-PFH \cite{ji2019invariant}        & 26.8 & 12.8 & 6.8 & 30.5 & 12.5 & 16.5 & 1.1 & 0.4 & 1.1 & 7.0 & 1.1 & 5.7 & 0.4 & 4.0 & 1.2 \\
& PICIE \cite{cho2021picie}            & 48.3 & 27.2 & 17.4 & 72.4 & 44.2 & 39.6 & 6.2 & 1.7 & 0.5 & 7.7 & 4.1 & 20.1 & 0.0 & 7.7 & 3.6 \\
& PICIE-S \cite{cho2021picie}          & 55.5 & 30.4 & 21.5 & 69.1 & 68.7 & 46.8 & 3.1 & 1.1 & 0.4 & 11.6 & 6.6 & 38.2 & 3.0 & 7.7 & 1.8 \\
& PICIE-PFH \cite{cho2021picie}        & 55.8 & 37.2 & 25.7 & 51.7 & 82.6 & 41.8 & 2.5 & 5.3 & 1.0 & 13.7 & 40.6 & 55.0 & 0.0 & 12.5 & 2.1 \\
& PICIE-S-PFH \cite{cho2021picie}      & 55.4 & 39.2 & 26.2 & 65.8 & 77.7 & 29.3 & 4.6 & 2.7 & 0.5 & 12.4 & 26.7 & 62.2 & 1.0 & 28.0 & 3.7 \\
& PointDC \cite{chen2023pointdc}          & 48.3 & 34.8 & 22.5 & 66.7 & 50.2 & 26.1 & 1.3 & 0.4 & 0.0 & 15.6 & 29.8 & 56.3 & 0.3 & 17.6 & 5.6 \\
& PointDC-DINOv2 \cite{chen2023pointdc}   & 77.1 & 50.3 & 38.1 & 90.8 & \textbf{92.0} & 57.3 & 10.2 & 0.6 & 34.2 & 20.3 & 46.4 & \textbf{85.6} & 0.0 & 19.1 & 0.9 \\
& GrowSP \cite{zhang2023growsp}           & 79.0 & 51.8 & 39.1 & 85.7 & 88.2 & 67.0 & \textbf{12.0} & \textbf{24.8} & 0.0 & 24.2 & 51.2 & 77.1 & 4.1 & 24.5 & 0.2 \\
& LogoSP \cite{zhang2025logosp}           & 77.0 & 52.2 & 39.4 & 92.3 & 67.7 & 72.0 & 7.2 & 0.7 & 44.8 & 32.4 & 58.0 & 53.3 & 0.2 & 44.1 & 0.5 \\
& GrowSP++ \cite{zhang2026growsp++}         & 79.5 & 54.5 & 43.8 & 90.9 & 91.2 & 66.7 & 8.9 & 0.4 & \textbf{67.5} & 43.7 & 57.0 & 54.4 & 0.3 & 43.8 & 0.8 \\
& \textbf{LangTail(ours)}              & \textbf{88.9} & \textbf{62.9} & \textbf{53.4} & \textbf{92.4} & 91.7 & \textbf{78.8} & 1.8 & 23.5 & 67.2 & \textbf{63.4} & \textbf{58.4} & 83.7 & \textbf{18.5} & \textbf{55.2} & 6.9 \\
\bottomrule
\end{tabular}
}
\end{table*}

\begin{table*}[h]
\centering
\caption{Quantitative results of our method and baselines on the Area-3 of S3DIS dataset.}
\label{S3DIS_area3}
\resizebox{\textwidth}{!}{ 
\begin{tabular}{ll ccccccccccccccc}
\toprule
\multicolumn{2}{l}{Method} & OA(\%) & mAcc(\%) & mIoU(\%) & ceil. & floor & wall & beam & col. & wind. & door & table & chair & sofa & book. & board. \\ 
\midrule
\multirow{3}{*}{\begin{tabular}[c]{@{}l@{}}Supervised \\ Methods\end{tabular}} 
& PointNet \cite{qi2017pointnet}      & 78.2 & 74.9 & 57.7 & 90.3 & 96.9 & 66.9 & 55.5 & 15.1 & 60.0 & 67.7 & 51.8 & 54.8 & 27.6 & 56.0 & 50.0 \\
& PointNet++ \cite{qi2017pointnet++}    & 79.8 & 85.9 & 65.8 & 91.4 & 98.0 & 68.5 & 50.1 & 15.2 & 74.8 & 74.7 & 63.2 & 70.1 & 53.6 & 54.0 & 76.5 \\
& SparseConv \cite{graham20183d}    & 91.3 & 86.8 & 78.6 & 93.1 & 96.2 & 80.4 & 74.7 & 63.3 & 77.2 & 69.5 & 80.1 & 85.5 & 89.5 & 80.1 & 52.2 \\
\midrule
\multirow{19}{*}{\begin{tabular}[c]{@{}l@{}}Unsupervised \\ Methods\end{tabular}} 
& RandCNN          & 25.3 & 21.7 & 10.8 & 25.5 & 32.6 & 17.2 & 4.9 & 3.3 & 10.6 & 9.1 & 4.2 & 2.9 & 0.9 & 9.3 & 8.8 \\
& van Kmeans       & 21.3 & 22.1 & 9.4 & 20.2 & 20.6 & 13.3 & 5.7 & 1.3 & 2.3 & 14.1 & 6.8 & 6.8 & 3.7 & 9.7 & 8.6 \\
& van Kmeans-S     & 21.2 & 21.9 & 9.3 & 19.8 & 18.8 & 14.4 & 5.0 & 1.6 & 1.5 & 14.4 & 7.4 & 6.6 & 4.2 & 10.3 & 7.7 \\
& van Kmeans-PFH   & 24.7 & 27.3 & 11.6 & 33.6 & 23.0 & 11.0 & 8.9 & 3.5 & 10.8 & 12.0 & 8.4 & 6.9 & 3.7 & 12.6 & 4.9 \\
& van Kmeans-S-PFH & 23.3 & 21.7 & 9.7 & 24.7 & 20.1 & 14.1 & 5.4 & 2.3 & 5.2 & 12.3 & 5.1 & 7.3 & 3.1 & 9.3 & 8.0 \\
& IIC \cite{ji2019invariant}             & 32.1 & 15.4 & 8.4 & 20.5 & 25.5 & 31.4 & 1.0 & 6.9 & 0.2 & 3.2 & 1.6 & 0.3 & 0.0 & 10.6 & 0.0 \\
& IIC-S \cite{ji2019invariant}            & 30.3 & 14.8 & 7.1 & 33.9 & 10.1 & 16.1 & 2.0 & 0.4 & 5.1 & 8.8 & 0.6 & 0.3 & 2.7 & 2.1 & 5.5 \\
& IIC-PFH \cite{ji2019invariant}          & 34.5 & 11.8 & 6.3 & 22.6 & 13.0 & 32.4 & 0.1 & 0.0 & 0.1 & 1.9 & 0.3 & 0.3 & 1.0 & 2.3 & 1.9 \\
& IIC-S-PFH \cite{ji2019invariant}        & 25.5 & 12.8 & 6.6 & 26.1 & 8.7 & 18.8 & 1.3 & 2.9 & 0.8 & 5.3 & 1.6 & 4.8 & 1.7 & 5.3 & 2.6 \\
& PICIE \cite{cho2021picie}            & 40.4 & 29.2 & 16.2 & 50.5 & 49.6 & 33.7 & 13.2 & 3.0 & 1.8 & 6.5 & 8.9 & 7.5 & 3.5 & 16.2 & 0.4 \\
& PICIE-S \cite{cho2021picie}          & 52.8 & 32.3 & 22.4 & 79.4 & 80.0 & 38.6 & 9.1 & 2.3 & 3.9 & 10.7 & 15.9 & 11.0 & 0.0 & 17.6 & 0.6 \\
& PICIE-PFH \cite{cho2021picie}        & 57.8 & 36.7 & 26.4 & 60.5 & 78.8 & 53.9 & 4.1 & 4.8 & 0.3 & 16.0 & 50.0 & 25.3 & 1.0 & 26.4 & 0.6 \\
& PICIE-S-PFH \cite{cho2021picie}      & 51.8 & 43.3 & 29.3 & 85.0 & 72.6 & 23.7 & 14.2 & 4.0 & 5.7 & 12.7 & 48.5 & 32.2 & 2.8 & 48.3 & 2.1 \\
& PointDC \cite{chen2023pointdc}          & 56.2 & 40.7 & 27.2 & 75.3 & 91.3 & 29.7 & 1.2 & 2.2 & 0.0 & 11.4 & 37.9 & 20.7 & 9.2 & 38.4 & 8.9 \\
& PointDC-DINOv2 \cite{chen2023pointdc}   & 70.5 & 52.4 & 39.3 & 86.2 & 93.4 & 48.8 & 0.0 & 14.5 & 38.6 & 28.2 & 68.7 & 55.0 & 0.0 & 37.7 & 0.0 \\
& GrowSP \cite{zhang2023growsp}           & 74.2 & 68.4 & 47.7 & \textbf{92.9} & 91.7 & 48.3 & \textbf{49.3} & 15.8 & 21.1 & 38.7 & 60.6 & 66.5 & 28.5 & 59.2 & 0.0 \\
& LogoSP \cite{zhang2025logosp}           & 79.8 & 62.7 & 48.9 & 90.0 & \textbf{94.3} & 65.9 & 16.0 & \textbf{18.6} & 67.8 & 45.5 & 59.2 & 56.3 & 3.6 & 69.4 & 0.0 \\
& GrowSP++ \cite{zhang2026growsp++}         & 78.7 & 65.6 & 50.2 & 92.3 & 94.0 & 59.3 & 30.6 & 15.5 & 68.5 & 53.1 & 60.7 & 56.2 & 0.6 & 71.2 & 0.0 \\
& \textbf{LangTail(ours)}              & \textbf{87.1} & \textbf{72.5} & \textbf{62.6} & 91.1 & 91.5 & \textbf{74.9} & 13.5 & 0.0 & \textbf{79.0} & \textbf{68.4} & \textbf{72.8} & \textbf{71.1} & \textbf{67.4} & \textbf{77.4} & \textbf{43.8} \\
\bottomrule
\end{tabular}
}
\end{table*}

\begin{table*}[h]
\centering
\caption{Quantitative results of our method and baselines on the Area-4 of S3DIS dataset.}
\label{tab:S3DIS_area4}
\resizebox{\textwidth}{!}{
\begin{tabular}{ll ccccccccccccccc}
\toprule
\multicolumn{2}{l}{Method} & OA(\%) & mAcc(\%) & mIoU(\%) & ceil. & floor & wall & beam & col. & wind. & door & table & chair & sofa & book. & board. \\ 
\midrule
\multirow{3}{*}{\begin{tabular}[c]{@{}l@{}}Supervised \\ Methods\end{tabular}} 
& PointNet \cite{qi2017pointnet}      & 73.0 & 58.6 & 41.6 & 81.3 & 95.7 & 68.4 & 1.3 & 22.4 & 29.0 & 44.8 & 39.3 & 42.5 & 17.6 & 36.6 & 20.1 \\
& PointNet++ \cite{qi2017pointnet++}    & 74.8 & 66.4 & 47.7 & 85.5 & 96.1 & 68.9 & 4.4 & 23.8 & 27.0 & 50.5 & 44.9 & 54.0 & 35.6 & 38.4 & 43.8 \\
& SparseConv \cite{graham20183d}    & 88.3 & 76.2 & 65.5 & 93.0 & 94.9 & 78.2 & 53.3 & 57.9 & 43.4 & 59.1 & 69.4 & 76.6 & 55.1 & 73.8 & 30.9 \\
\midrule
\multirow{19}{*}{\begin{tabular}[c]{@{}l@{}}Unsupervised \\ Methods\end{tabular}} 
& RandCNN          & 22.6 & 18.3 & 9.0  & 22.0 & 28.2 & 16.4 & 0.4 & 4.2 & 5.6 & 11.9 & 4.5 & 4.3 & 1.4 & 6.4 & 2.3 \\
& van Kmeans       & 17.9 & 19.9 & 7.8  & 18.6 & 18.2 & 10.6 & 0.9 & 3.8 & 5.2 & 11.7 & 5.8 & 7.4 & 2.4 & 8.7 & 0.4 \\
& van Kmeans-S     & 17.3 & 19.6 & 7.4  & 16.7 & 14.8 & 10.2 & 0.9 & 4.4 & 5.1 & 10.5 & 6.1 & 7.5 & 2.8 & 9.8 & 0.0 \\
& van Kmeans-PFH   & 21.7 & 22.3 & 9.3  & 28.9 & 21.3 & 12.6 & 1.2 & 5.9 & 4.4 & 11.1 & 6.6 & 8.6 & 2.6 & 7.4 & 0.7 \\
& van Kmeans-S-PFH & 23.1 & 18.3 & 8.9  & 24.4 & 26.2 & 14.5 & 1.7 & 5.0 & 3.2 & 11.7 & 3.8 & 6.1 & 0.4 & 9.0 & 1.1 \\
& IIC \cite{ji2019invariant}              & 33.0 & 13.5 & 8.2  & 14.9 & 25.9 & 35.1 & 0.0 & 1.1 & 1.4 & 3.7  & 4.1 & 0.9 & 0.0 & 10.8 & 0.0 \\
& IIC-S \cite{ji2019invariant}            & 26.7 & 13.5 & 6.1  & 27.6 & 4.9  & 15.3 & 0.1 & 1.0 & 5.3 & 12.0 & 1.5 & 2.3 & 1.1 & 1.6 & 0.0 \\
& IIC-PFH \cite{ji2019invariant}          & 31.0 & 11.4 & 6.3  & 21.0 & 10.9 & 30.0 & 0.1 & 1.3 & 3.0 & 2.3  & 0.8 & 1.3 & 0.0 & 4.6 & 0.1 \\
& IIC-S-PFH \cite{ji2019invariant}        & 27.3 & 13.2 & 6.9  & 29.4 & 9.9  & 20.0 & 0.1 & 2.6 & 3.1 & 4.0  & 3.7 & 2.5 & 0.6 & 6.3 & 0.8 \\
& PICIE \cite{cho2021picie}            & 43.2 & 29.4 & 17.8 & 62.2 & 72.7 & 22.6 & 2.5 & 3.4 & 3.5 & 8.8  & 4.1 & 17.4 & 0.0 & 15.5 & 0.7 \\
& PICIE-S \cite{cho2021picie}          & 50.0 & 32.2 & 22.3 & 72.7 & 77.7 & 37.4 & 1.6 & 2.4 & 3.3 & 11.3 & 25.8 & 19.7 & 3.5 & 12.3 & 0.0 \\
& PICIE-PFH \cite{cho2021picie}        & 58.6 & 42.6 & 28.5 & 69.5 & 83.6 & 48.3 & 2.6 & 9.4 & 5.8 & 16.4 & 44.1 & 36.3 & 1.6 & 23.9 & 0.1 \\
& PICIE-S-PFH \cite{cho2021picie}      & 47.1 & 44.8 & 27.6 & 67.2 & 67.8 & 25.9 & 2.6 & 14.5 & 5.2 & 10.2 & 47.9 & 44.1 & 0.9 & 44.2 & 0.4 \\
& PointDC \cite{chen2023pointdc}          & 54.0 & 35.3 & 25.2 & 87.9 & 86.8 & 24.7 & 0.0 & 4.2 & 12.2 & 18.7 & 32.6 & 17.6 & 1.9 & 15.6 & 0.0 \\
& PointDC-DINOv2 \cite{chen2023pointdc}   & 73.0 & 46.5 & 39.8 & 89.4 & 91.9 & 58.2 & 0.0 & 0.7 & 25.3 & 19.0 & 52.4 & 53.8 & 0.7 & 59.0 & 0.0 \\
& GrowSP \cite{zhang2023growsp}           & 76.0 & 59.8 & 42.8 & 90.6 & 91.5 & 64.4 & \textbf{15.9} & 7.6 & 27.4 & 31.5 & 52.0 & 67.4 & 16.8 & 48.5 & 0.0 \\
& LogoSP \cite{zhang2025logosp}           & 80.8 & 54.4 & 43.5 & \textbf{92.5} & \textbf{93.8} & 73.9 & 4.1 & 0.1 & 34.8 & 44.6 & 54.0 & 64.9 & \textbf{69.2} & 73.4 & \textbf{33.5} \\
& GrowSP++ \cite{zhang2026growsp++}         & 77.3 & 52.6 & 42.8 & 91.0 & 92.3 & 65.0 & 0.0 & 6.4 & 42.0 & 47.1 & 52.7 & 60.7 & 0.5 & 55.2 & 0.0 \\
& \textbf{LangTail(ours)}              & \textbf{87.6} & \textbf{70.6} & \textbf{59.5} & 90.3 & 93.3 & \textbf{76.9} & 0.0 & \textbf{39.4} & \textbf{50.7} & \textbf{63.6} & \textbf{72.5} & \textbf{74.9} & 55.8 & \textbf{76.1} & 21.2 \\
\bottomrule
\end{tabular}
}
\end{table*}

\begin{table*}[h]
\centering
\caption{Quantitative results of our method and baselines on the Area-5 of S3DIS dataset.}
\label{tab:S3DIS_area5}
\resizebox{\textwidth}{!}{
\begin{tabular}{ll ccccccccccccccc}
\toprule
\multicolumn{2}{l}{Method} & OA(\%) & mAcc(\%) & mIoU(\%) & ceil. & floor & wall & beam & col. & wind. & door & table & chair & sofa & book. & board. \\ 
\midrule
\multirow{3}{*}{\begin{tabular}[c]{@{}l@{}}Supervised \\ Methods\end{tabular}} 
& PointNet \cite{qi2017pointnet}      & 77.5 & 59.1 & 44.6 & 85.2 & 97.4 & 72.3 & 0.1 & 10.6 & 54.9 & 18.5 & 48.4 & 39.5 & 12.4 & 55.5 & 40.2 \\
& PointNet++ \cite{qi2017pointnet++}    & 77.5 & 62.6 & 50.1 & 83.1 & 97.2 & 66.4 & 0.0 & 8.1 & 55.6 & 15.2 & 60.4 & 64.5 & 36.6 & 58.3 & 55.7 \\
& SparseConv \cite{graham20183d}    & 88.4 & 69.2 & 60.8 & 92.6 & 95.9 & 77.2 & 0.1 & 36.7 & 37.6 & 59.8 & 77.2 & 83.9 & 59.7 & 78.5 & 30.39 \\
\midrule
\multirow{19}{*}{\begin{tabular}[c]{@{}l@{}}Unsupervised \\ Methods\end{tabular}} 
& RandCNN          & 23.3 & 17.3 & 9.2  & 25.3 & 24.5 & 17.4 & 0.0 & 2.3  & 12.5 & 6.5  & 5.7  & 3.0  & 0.3 & 10.1 & 2.2 \\
& van Kmeans       & 21.4 & 21.2 & 8.7  & 18.7 & 18.0 & 16.7 & 0.2 & 2.5  & 12.0 & 5.7  & 8.7  & 5.6  & 0.0 & 13.6 & 2.3 \\
& van Kmeans-S     & 21.9 & 22.9 & 9.0  & 19.3 & 18.1 & 17.0 & 0.2 & 2.1  & 11.8 & 4.5  & 8.9  & 6.6  & 0.2 & 14.0 & 4.8 \\
& van Kmeans-PFH   & 23.2 & 23.6 & 10.2 & 32.0 & 20.5 & 10.3 & 0.1 & 3.6  & 15.2 & 7.1  & 9.9  & 6.2  & 0.7 & 12.4 & 4.9 \\
& van Kmeans-S-PFH & 22.8 & 20.6 & 9.2  & 25.2 & 26.5 & 12.7 & 0.4 & 2.0  & 8.7  & 8.1  & 5.8  & 5.7  & 0.0 & 12.7 & 3.3 \\
& IIC \cite{ji2019invariant}              & 28.5 & 12.5 & 6.4  & 6.1  & 19.8 & 27.9 & 0.0 & 2.1  & 0.1  & 3.4  & 7.9  & 0.4  & 0.0 & 8.6  & 0.0 \\
& IIC-S \cite{ji2019invariant}            & 29.2 & 13.0 & 6.8  & 28.9 & 12.3 & 18.7 & 0.0 & 0.1  & 3.6  & 1.3  & 3.8  & 0.6  & 0.0 & 8.1  & 3.8 \\
& IIC-PFH \cite{ji2019invariant}          & 28.6 & 16.8 & 7.9  & 23.7 & 24.9 & 17.7 & 5.9 & 1.4  & 12.6 & 0.2  & 5.3  & 0.6  & 0.0 & 2.4  & 0.0 \\
& IIC-S-PFH \cite{ji2019invariant}        & 31.2 & 16.3 & 9.1  & 43.2 & 23.6 & 14.9 & 0.0 & 1.6  & 3.9  & 2.4  & 3.6  & 1.5  & 0.8 & 9.5  & 4.5 \\
& PICIE \cite{cho2021picie}            & 61.6 & 25.8 & 17.9 & 65.7 & 61.4 & 58.4 & 0.0 & 0.3  & 2.2  & 1.7  & 12.1 & 0.0  & 0.0 & 12.4 & 0.0 \\
& PICIE-S \cite{cho2021picie}          & 49.6 & 28.9 & 20.0 & 64.2 & 75.1 & 42.4 & 0.1 & 1.2  & 4.6  & 7.4  & 18.7 & 9.2  & 1.0 & 16.0 & 0.4 \\
& PICIE-PFH \cite{cho2021picie}        & 54.0 & 36.8 & 24.4 & 58.4 & 68.6 & 49.9 & 0.1 & 7.6  & 5.3  & 14.2 & 45.1 & 16.3 & 0.1 & 27.1 & 0.6 \\
& PICIE-S-PFH \cite{cho2021picie}      & 48.4 & 40.4 & 25.2 & 59.6 & 72.5 & 26.0 & 0.2 & 8.5  & 5.9  & 8.7  & 46.0 & 26.9 & 0.4 & 46.8 & 0.3 \\
& PointDC \cite{chen2023pointdc}          & 55.5 & 35.1 & 23.9 & 84.4 & 84.3 & 30.2 & 0.0 & 1.8  & 12.2 & 7.1  & 24.6 & 6.9  & 5.4 & 29.7 & 0.7 \\
& PointDC-DINOv2 \cite{chen2023pointdc}   & 75.7 & 48.7 & 40.2 & 87.7 & 89.5 & 59.2 & 0.0 & 0.8  & 25.8 & 26.3 & 62.0 & 68.3 & 1.5 & 61.0 & 0.5 \\
& GrowSP \cite{zhang2023growsp}           & 78.4 & 57.2 & 44.5 & 90.5 & 90.1 & 66.7 & 0.0 & \textbf{14.8} & 27.6 & \textbf{45.6} & 59.4 & 71.9 & 10.7 & 56.0 & 0.2 \\
& LogoSP \cite{zhang2025logosp}           & 82.8 & 55.9 & 46.5 & \textbf{92.9} & \textbf{95.4} & 73.2 & 0.0 & 3.3  & 57.8 & 35.9 & 55.5 & \textbf{74.6} & 1.9 & 67.3 & 0.3 \\
& GrowSP++ \cite{zhang2026growsp++}         & 78.7 & 60.1 & 46.6 & 87.3 & 92.8 & 53.5 & 0.0 & 5.3  & 54.2 & 24.4 & 56.8 & 63.4 & 52.1 & 69.0 & 0.0 \\
&\textbf{LangTail(ours)}             & \textbf{86.7} & \textbf{69.6} & \textbf{59.5} & 92.5 & 90.5 & \textbf{73.8} & \textbf{8.2} & 8.7  & \textbf{58.0} & 38.0 & \textbf{73.3} & 71.0 & \textbf{68.7} & \textbf{76.7} & \textbf{54.9} \\
\bottomrule
\end{tabular}
}
\end{table*}

\begin{table*}[t]
\centering
\caption{Quantitative results of our method and baselines on the Area-6 of S3DIS dataset.}
\label{tab:S3DIS_area6}
\resizebox{\textwidth}{!}{ 
\begin{tabular}{ll ccccccccccccccc}
\toprule
\multicolumn{2}{l}{Method} & OA(\%) & mAcc(\%) & mIoU(\%) & ceil. & floor & wall & beam & col. & wind. & door & table & chair & sofa & book. & board. \\ 
\midrule
\multirow{3}{*}{\begin{tabular}[c]{@{}l@{}}Supervised \\ Methods\end{tabular}} 
& PointNet \cite{qi2017pointnet}      & 79.0 & 79.6 & 60.9 & 85.7 & 96.5 & 71.8 & 59.4 & 47.4 & 67.4 & 74.3 & 56.2 & 48.9 & 20.9 & 50.0 & 52.5 \\
& PointNet++ \cite{qi2017pointnet++}    & 82.0 & 89.3 & 69.0 & 87.5 & 96.3 & 76.8 & 66.4 & 54.4 & 72.1 & 77.4 & 64.3 & 66.5 & 43.7 & 51.8 & 70.2 \\
& SparseConv \cite{graham20183d}    & 91.6 & 87.3 & 80.5 & 97.4 & 95.0 & 83.4 & 83.0 & 75.1 & 81.1 & 74.9 & 81.3 & 84.3 & 79.0 & 80.7 & 61.4 \\
\midrule
\multirow{19}{*}{\begin{tabular}[c]{@{}l@{}}Unsupervised \\ Methods\end{tabular}} 
& RandCNN          & 22.1 & 16.0 & 8.5  & 17.6 & 24.2 & 19.2 & 0.0  & 1.7  & 12.2 & 7.6  & 6.2  & 2.6  & 0.2  & 8.9  & 1.7  \\
& van Kmeans       & 21.0 & 25.0 & 10.4 & 18.6 & 17.6 & 8.9  & 11.3 & 0.6  & 14.8 & 17.6 & 12.0 & 8.7  & 0.3  & 7.8  & 6.2  \\
& van Kmeans-S     & 20.6 & 26.3 & 10.2 & 16.6 & 17.8 & 9.0  & 13.4 & 0.6  & 15.0 & 17.0 & 10.7 & 8.3  & 1.8  & 7.4  & 5.1  \\
& van Kmeans-PFH   & 25.8 & 27.2 & 12.8 & 32.0 & 25.1 & 12.2 & 15.0 & 3.6  & 19.7 & 14.9 & 9.4  & 10.9 & 2.1  & 5.3  & 3.5  \\
& van Kmeans-S-PFH & 23.9 & 23.0 & 10.5 & 22.2 & 20.8 & 15.2 & 14.9 & 0.3  & 9.5  & 13.6 & 10.9 & 7.4  & 1.6  & 5.0  & 4.9  \\
& IIC \cite{ji2019invariant}              & 32.5 & 15.9 & 9.2  & 21.9 & 33.8 & 29.1 & 3.1  & 15.2 & 0.0  & 2.7  & 0.7  & 0.0  & 0.0  & 1.5  & 1.8  \\
& IIC-S \cite{ji2019invariant}            & 27.0 & 16.5 & 7.6  & 28.9 & 8.5  & 12.7 & 0.5  & 0.2  & 0.0  & 14.1 & 10.4 & 0.6  & 0.0  & 1.0  & 3.8  \\
& IIC-PFH \cite{ji2019invariant}          & 28.4 & 16.7 & 7.8  & 23.6 & 24.5 & 17.4 & 5.8  & 1.5  & 12.6 & 0.2  & 4.8  & 0.6  & 0.0  & 2.4  & 0.0  \\
& IIC-S-PFH \cite{ji2019invariant}        & 22.9 & 13.1 & 6.7  & 28.8 & 9.0  & 12.6 & 0.6  & 4.8  & 2.1  & 6.6  & 4.4  & 1.2  & 0.3  & 5.4  & 4.9  \\
& PICIE \cite{cho2021picie}            & 39.3 & 28.5 & 17.8 & 56.9 & 61.7 & 18.6 & 20.5 & 4.2  & 6.0  & 8.7  & 14.7 & 15.9 & 1.1  & 5.7  & 0.0  \\
& PICIE-S \cite{cho2021picie}          & 47.4 & 30.7 & 21.8 & 65.5 & 67.5 & 37.2 & 16.6 & 2.0  & 4.8  & 10.6 & 23.7 & 23.4 & 23.4 & 1.7  & 0.0  \\
& PICIE-PFH \cite{cho2021picie}        & 51.8 & 41.3 & 27.9 & 63.1 & 56.5 & 39.0 & 11.4 & 10.0 & 5.3  & 19.1 & 63.8 & 50.4 & 0.5  & 37.2 & 0.9  \\
& PICIE-S-PFH \cite{cho2021picie}      & 44.0 & 36.1 & 24.7 & 58.2 & 60.4 & 24.5 & 17.9 & 10.1 & 8.1  & 12.7 & 44.0 & 5.5  & 0.5  & 54.3 & 0.3  \\
& PointDC \cite{chen2023pointdc}          & 62.4 & 38.7 & 28.6 & 85.8 & 85.6 & 43.8 & 5.0  & 16.5 & 8.8  & 10.7 & 41.7 & 12.5 & 0.0  & 33.0 & 0.0  \\
& PointDC-DINOv2 \cite{chen2023pointdc}   & 76.4 & 55.5 & 46.4 & 90.3 & 91.4 & 61.7 & 0.0  & 19.7 & 63.1 & 33.6 & 67.9 & 65.7 & 1.4  & 62.5 & 0.0  \\
& GrowSP \cite{zhang2023growsp}           & 75.6 & 58.5 & 47.6 & 89.4 & 88.0 & 57.7 & \textbf{70.6} & 2.0  & 32.4 & 36.7 & 63.2 & 69.8 & 1.5  & 58.9 & 0.2  \\
& LogoSP \cite{zhang2025logosp}           & 77.9 & 62.9 & 50.6 & \textbf{94.6} & \textbf{92.7} & 64.0 & 25.7 & 23.9 & 65.9 & 38.8 & 68.9 & 72.2 & 2.5  & 68.4 & 0.0  \\
& GrowSP++ \cite{zhang2026growsp++}         & 78.4 & 63.5 & 51.5 & 86.8 & 92.6 & 63.9 & 36.4 & 21.2 & \textbf{73.4} & 39.8 & 66.6 & 71.8 & 1.8  & 64.7 & 0.0  \\
& \textbf{LangTail(ours)}              & \textbf{85.1} & \textbf{82.4} & \textbf{67.2} & 88.7 & 88.3 & \textbf{69.0} & 47.7 & \textbf{45.5} & 68.9 & \textbf{73.0} & \textbf{75.5} & \textbf{72.4} & \textbf{63.4} & \textbf{71.5} & \textbf{42.4} \\
\bottomrule
\end{tabular}
}
\end{table*}

\paragraph{nuScenes.}
The table~\ref{tab:nuScenes_per} shows the detailed results for each category. Our results outperform all baselines, demonstrating that our method is effective on outdoor datasets as well.
\begin{table*}[t]
\centering
\caption{Per-category quantitative results on the validation split of nuScenes dataset.}
\label{tab:nuScenes_per}
\resizebox{\textwidth}{!}{
\begin{tabular}{l ccccccccccccccccccc}
\toprule
Method & OA(\%) & mAcc(\%) & mIoU(\%) & barrier & bicycle & bus & car & cons. veh. & motor. & ped. & cone & trailer & truck & driv. surf. & other flat & sidewalk & terrain & manmade & veget. \\ 
\midrule
GrowSP \cite{zhang2023growsp}         & 39.2 & 17.5 & 10.2 & 7.5  & 0.0 & 0.4  & 42.9 & 0.1 & 0.0 & 0.6  & 0.0 & 0.7  & 1.4  & 48.4 & 0.8 & 6.5 & 13.1 & 21.4 & 19.7 \\
PointDC \cite{chen2023pointdc}        & 56.8 & 29.4 & 17.7 & 11.6 & 0.0 & 0.5  & 63.1 & 0.3 & 0.0 & 4.4  & 0.0 & 1.2  & 26.4 & 70.1 & 0.1 & 7.1 & 19.3 & 21.1 & 58.1 \\
PointDC-DINOv2 \cite{chen2023pointdc} & 51.8 & 28.6 & 18.2 & \textbf{17.0} & 0.0 & 0.2  & 58.4 & 0.2 & 0.0 & 1.5  & 0.0 & 1.6  & 43.3 & \textbf{71.8} & 0.0 & \textbf{8.3} & 19.5 & 17.6 & 51.8 \\
LogoSP \cite{zhang2025logosp}         & 54.8 & 29.2 & 20.1 & 16.6 & 0.0 & 0.7  & 70.2 & 0.2 & 0.2 & \textbf{33.6} & 0.0 & 0.3  & 38.4 & 59.4 & 0.4 & 8.0 & 10.7 & 33.0 & 49.3 \\
GrowSP++ \cite{zhang2026growsp++}       & 55.7 & 30.4 & 19.7 & 16.5 & 0.0 & 33.8 & 73.0 & \textbf{0.4} & 0.0 & 0.8  & 0.0 & 1.0  & 44.7 & 63.5 & \textbf{1.1} & 3.6 & 12.7 & 20.6 & 49.9 \\
Ours           & \textbf{73.9} & \textbf{35.1} & \textbf{29.0} & 6.0  & 0.0 & \textbf{50.2} & \textbf{74.0} & 0.0 & \textbf{5.0} & 26.3 & 0.0 & \textbf{13.7} & \textbf{51.6} & 68.4 & 0.0 & 6.1 & \textbf{23.7} & \textbf{71.7} & \textbf{66.9} \\
\bottomrule
\end{tabular}
}
\end{table*}

\section{Limitations}
\label{sec:limitations}
While langTail achieves outperforming results across multiple benchmarks, several limitations remain.
\paragraph{Limited 2D coverage in outdoor scenes.}
langTail relies on multi-view RGB images to distill semantic priors from 2D to 3D via DINOv2. However, outdoor LiDAR datasets often provide sparse or temporally misaligned image coverage with narrow fields of view, resulting in insufficient 2D-3D correspondences for effective distillation. 
\paragraph{Sensitivity to pseudo-label quality.}
The entity branch leverages a pre-generated vocabulary to guide semantic discovery and pseudo-label assignment. Its stability is therefore contingent on the coverage, precision, and scene-relevance of this vocabulary. If the generation process yields redundant, ambiguous, or domain-mismatched terms, cross-modal alignment may propagate noise into the clustering pipeline.


\clearpage

\end{document}

%% file: math_commands.tex
\usepackage{amsmath,amsfonts,bm}

\def\eqref#1{equation~\ref{#1}}

\def\1{\bm{1}}

\def\vb{{\bm{b}}}

\def\vf{{\bm{f}}}

\def\vu{{\bm{u}}}
\def\vv{{\bm{v}}}

\def\mA{{\bm{A}}}
\def\mB{{\bm{B}}}

\def\mD{{\bm{D}}}

\def\mF{{\bm{F}}}
\def\mG{{\bm{G}}}

\def\mI{{\bm{I}}}

\def\mL{{\bm{L}}}
\def\mM{{\bm{M}}}

\def\mP{{\bm{P}}}

\def\mU{{\bm{U}}}
\def\mV{{\bm{V}}}

\def\mX{{\bm{X}}}

\def\mLambda{{\bm{\Lambda}}}

\DeclareMathAlphabet{\mathsfit}{\encodingdefault}{\sfdefault}{m}{sl}
\SetMathAlphabet{\mathsfit}{bold}{\encodingdefault}{\sfdefault}{bx}{n}

\def\gG{{\mathcal{G}}}

\def\gK{{\mathcal{K}}}

\def\sR{{\mathbb{R}}}

\newcommand{\Ls}{\mathcal{L}}



%% file: reference.bib
@inproceedings{zhang2023growsp,
  title={Growsp: Unsupervised semantic segmentation of 3d point clouds},
  author={Zhang, Zihui and Yang, Bo and Wang, Bing and Li, Bo},
  booktitle={Proceedings of the IEEE/CVF Conference on computer vision and pattern recognition},
  pages={17619--17629},
  year={2023}
}

@inproceedings{chen2023pointdc,
  title={PointDC: Unsupervised semantic segmentation of 3D point clouds via cross-modal distillation and super-voxel clustering},
  author={Chen, Zisheng and Xu, Hongbin and Chen, Weitao and Zhou, Zhipeng and Xiao, Haihong and Sun, Baigui and Xie, Xuansong and others},
  booktitle={Proceedings of the IEEE/CVF International Conference on Computer Vision},
  pages={14290--14299},
  year={2023}
}

@inproceedings{zhang2025logosp,
  title={Logosp: Local-global grouping of superpoints for unsupervised semantic segmentation of 3d point clouds},
  author={Zhang, Zihui and Dai, Weisheng and Wen, Hongtao and Yang, Bo},
  booktitle={Proceedings of the IEEE/CVF Conference on Computer Vision and Pattern Recognition},
  pages={1374--1384},
  year={2025}
}

@inproceedings{liu2024u3ds3,
  title={U3ds3: Unsupervised 3d semantic scene segmentation},
  author={Liu, Jiaxu and Yu, Zhengdi and Breckon, Toby P and Shum, Hubert PH},
  booktitle={Proceedings of the IEEE/CVF Winter Conference on Applications of Computer Vision},
  pages={3759--3768},
  year={2024}
}

@inproceedings{cho2021picie,
  title={Picie: Unsupervised semantic segmentation using invariance and equivariance in clustering},
  author={Cho, Jang Hyun and Mall, Utkarsh and Bala, Kavita and Hariharan, Bharath},
  booktitle={Proceedings of the IEEE/CVF conference on computer vision and pattern recognition},
  pages={16794--16804},
  year={2021}
}

@inproceedings{ji2019invariant,
  title={Invariant information clustering for unsupervised image classification and segmentation},
  author={Ji, Xu and Henriques, Joao F and Vedaldi, Andrea},
  booktitle={Proceedings of the IEEE/CVF international conference on computer vision},
  pages={9865--9874},
  year={2019}
}

@inproceedings{caron2018deep,
  title={Deep clustering for unsupervised learning of visual features},
  author={Caron, Mathilde and Bojanowski, Piotr and Joulin, Armand and Douze, Matthijs},
  booktitle={Proceedings of the European conference on computer vision (ECCV)},
  pages={132--149},
  year={2018}
}

@inproceedings{xie2020pointcontrast,
  title={Pointcontrast: Unsupervised pre-training for 3d point cloud understanding},
  author={Xie, Saining and Gu, Jiatao and Guo, Demi and Qi, Charles R and Guibas, Leonidas and Litany, Or},
  booktitle={European conference on computer vision},
  pages={574--591},
  year={2020},
  organization={Springer}
}

@inproceedings{hou2021exploring,
  title={Exploring data-efficient 3d scene understanding with contrastive scene contexts},
  author={Hou, Ji and Graham, Benjamin and Nie{\ss}ner, Matthias and Xie, Saining},
  booktitle={Proceedings of the IEEE/CVF conference on computer vision and pattern recognition},
  pages={15587--15597},
  year={2021}
}

@article{zhang2026growsp++,
  title={GrowSP++: Growing Superpoints and Primitives for Unsupervised 3D Semantic Segmentation},
  author={Zhang, Zihui and Dai, Weisheng and Wang, Bing and Li, Bo and Yang, Bo},
  journal={IEEE Transactions on Pattern Analysis and Machine Intelligence},
  year={2026},
  publisher={IEEE}
}

@inproceedings{zhan2026p,
  title={P-SLCR: Unsupervised Point Cloud Semantic Segmentation via Prototypes Structure Learning and Consistent Reasoning},
  author={Zhan, Lixin and Jie, Jiang and Zhou, Tianjian and Du, Yukun and Zheng, Yan and Duan, Xuehu},
  booktitle={Proceedings of the AAAI Conference on Artificial Intelligence},
  volume={40},
  number={15},
  pages={12349--12357},
  year={2026}
}

@inproceedings{van2020scan,
  title={Scan: Learning to classify images without labels},
  author={Van Gansbeke, Wouter and Vandenhende, Simon and Georgoulis, Stamatios and Proesmans, Marc and Van Gool, Luc},
  booktitle={European conference on computer vision},
  pages={268--285},
  year={2020},
  organization={Springer}
}

@article{hamilton2022unsupervised,
  title={Unsupervised semantic segmentation by distilling feature correspondences},
  author={Hamilton, Mark and Zhang, Zhoutong and Hariharan, Bharath and Snavely, Noah and Freeman, William T},
  journal={arXiv preprint arXiv:2203.08414},
  year={2022}
}

@inproceedings{radford2021learning,
  title={Learning transferable visual models from natural language supervision},
  author={Radford, Alec and Kim, Jong Wook and Hallacy, Chris and Ramesh, Aditya and Goh, Gabriel and Agarwal, Sandhini and Sastry, Girish and Askell, Amanda and Mishkin, Pamela and Clark, Jack and others},
  booktitle={International conference on machine learning},
  pages={8748--8763},
  year={2021},
  organization={PmLR}
}

@inproceedings{zhu2023pointclip,
  title={Pointclip v2: Prompting clip and gpt for powerful 3d open-world learning},
  author={Zhu, Xiangyang and Zhang, Renrui and He, Bowei and Guo, Ziyu and Zeng, Ziyao and Qin, Zipeng and Zhang, Shanghang and Gao, Peng},
  booktitle={Proceedings of the IEEE/CVF international conference on computer vision},
  pages={2639--2650},
  year={2023}
}

@inproceedings{peng2023openscene,
  title={Openscene: 3d scene understanding with open vocabularies},
  author={Peng, Songyou and Genova, Kyle and Jiang, Chiyu and Tagliasacchi, Andrea and Pollefeys, Marc and Funkhouser, Thomas and others},
  booktitle={Proceedings of the IEEE/CVF conference on computer vision and pattern recognition},
  pages={815--824},
  year={2023}
}

@inproceedings{xue2023ulip,
  title={Ulip: Learning a unified representation of language, images, and point clouds for 3d understanding},
  author={Xue, Le and Gao, Mingfei and Xing, Chen and Mart{\'\i}n-Mart{\'\i}n, Roberto and Wu, Jiajun and Xiong, Caiming and Xu, Ran and Niebles, Juan Carlos and Savarese, Silvio},
  booktitle={Proceedings of the IEEE/CVF conference on computer vision and pattern recognition},
  pages={1179--1189},
  year={2023}
}

@article{liu2023openshape,
  title={Openshape: Scaling up 3d shape representation towards open-world understanding},
  author={Liu, Minghua and Shi, Ruoxi and Kuang, Kaiming and Zhu, Yinhao and Li, Xuanlin and Han, Shizhong and Cai, Hong and Porikli, Fatih and Su, Hao},
  journal={Advances in neural information processing systems},
  volume={36},
  pages={44860--44879},
  year={2023}
}

@article{thengane2025foundational,
  title={Foundational models for 3d point clouds: A survey and outlook},
  author={Thengane, Vishal and Zhu, Xiatian and Bouzerdoum, Salim and Phung, Son Lam and Li, Yunpeng},
  journal={arXiv preprint arXiv:2501.18594},
  year={2025}
}

@inproceedings{yang2024regionplc,
  title={Regionplc: Regional point-language contrastive learning for open-world 3d scene understanding},
  author={Yang, Jihan and Ding, Runyu and Deng, Weipeng and Wang, Zhe and Qi, Xiaojuan},
  booktitle={Proceedings of the IEEE/CVF conference on computer vision and pattern recognition},
  pages={19823--19832},
  year={2024}
}

@article{krusi2017driving,
  title={Driving on point clouds: Motion planning, trajectory optimization, and terrain assessment in generic nonplanar environments},
  author={Kr{\"u}si, Philipp and Furgale, Paul and Bosse, Michael and Siegwart, Roland},
  journal={Journal of Field Robotics},
  volume={34},
  number={5},
  pages={940--984},
  year={2017},
  publisher={Wiley Online Library}
}

@article{ze20243d,
  title={3d diffusion policy: Generalizable visuomotor policy learning via simple 3d representations},
  author={Ze, Yanjie and Zhang, Gu and Zhang, Kangning and Hu, Chenyuan and Wang, Muhan and Xu, Huazhe},
  journal={arXiv preprint arXiv:2403.03954},
  year={2024}
}

@article{huang2026pointworld,
  title={PointWorld: Scaling 3D World Models for In-The-Wild Robotic Manipulation},
  author={Huang, Wenlong and Chao, Yu-Wei and Mousavian, Arsalan and Liu, Ming-Yu and Fox, Dieter and Mo, Kaichun and Fei-Fei, Li},
  journal={arXiv preprint arXiv:2601.03782},
  year={2026}
}

@article{cui2021deep,
  title={Deep learning for image and point cloud fusion in autonomous driving: A review},
  author={Cui, Yaodong and Chen, Ren and Chu, Wenbo and Chen, Long and Tian, Daxin and Li, Ying and Cao, Dongpu},
  journal={IEEE Transactions on Intelligent Transportation Systems},
  volume={23},
  number={2},
  pages={722--739},
  year={2021},
  publisher={IEEE}
}

@inproceedings{hu2020randla,
  title={Randla-net: Efficient semantic segmentation of large-scale point clouds},
  author={Hu, Qingyong and Yang, Bo and Xie, Linhai and Rosa, Stefano and Guo, Yulan and Wang, Zhihua and Trigoni, Niki and Markham, Andrew},
  booktitle={Proceedings of the IEEE/CVF conference on computer vision and pattern recognition},
  pages={11108--11117},
  year={2020}
}

@inproceedings{wu2024point,
  title={Point transformer v3: Simpler faster stronger},
  author={Wu, Xiaoyang and Jiang, Li and Wang, Peng-Shuai and Liu, Zhijian and Liu, Xihui and Qiao, Yu and Ouyang, Wanli and He, Tong and Zhao, Hengshuang},
  booktitle={Proceedings of the IEEE/CVF conference on computer vision and pattern recognition},
  pages={4840--4851},
  year={2024}
}

@inproceedings{kolodiazhnyi2024oneformer3d,
  title={Oneformer3d: One transformer for unified point cloud segmentation},
  author={Kolodiazhnyi, Maxim and Vorontsova, Anna and Konushin, Anton and Rukhovich, Danila},
  booktitle={Proceedings of the IEEE/CVF Conference on Computer Vision and Pattern Recognition},
  pages={20943--20953},
  year={2024}
}

@article{bai2025qwen3,
  title={Qwen3-vl technical report},
  author={Bai, Shuai and Cai, Yuxuan and Chen, Ruizhe and Chen, Keqin and Chen, Xionghui and Cheng, Zesen and Deng, Lianghao and Ding, Wei and Gao, Chang and Ge, Chunjiang and others},
  journal={arXiv preprint arXiv:2511.21631},
  year={2025}
}

@article{yang2025qwen3,
  title={Qwen3 technical report},
  author={Yang, An and Li, Anfeng and Yang, Baosong and Zhang, Beichen and Hui, Binyuan and Zheng, Bo and Yu, Bowen and Gao, Chang and Huang, Chengen and Lv, Chenxu and others},
  journal={arXiv preprint arXiv:2505.09388},
  year={2025}
}

@article{guo2025seed1,
  title={Seed1. 5-vl technical report},
  author={Guo, Dong and Wu, Faming and Zhu, Feida and Leng, Fuxing and Shi, Guang and Chen, Haobin and Fan, Haoqi and Wang, Jian and Jiang, Jianyu and Wang, Jiawei and others},
  journal={arXiv preprint arXiv:2505.07062},
  year={2025}
}

@article{seed2026seed1,
  title={Seed1. 8 model card: Towards generalized real-world agency},
  author={Seed, Bytedance},
  journal={arXiv preprint arXiv:2603.20633},
  year={2026}
}

@inproceedings{behley2019semantickitti,
  title={Semantickitti: A dataset for semantic scene understanding of lidar sequences},
  author={Behley, Jens and Garbade, Martin and Milioto, Andres and Quenzel, Jan and Behnke, Sven and Stachniss, Cyrill and Gall, Jurgen},
  booktitle={Proceedings of the IEEE/CVF international conference on computer vision},
  pages={9297--9307},
  year={2019}
}

@inproceedings{hu2021towards,
  title={Towards semantic segmentation of urban-scale 3D point clouds: A dataset, benchmarks and challenges},
  author={Hu, Qingyong and Yang, Bo and Khalid, Sheikh and Xiao, Wen and Trigoni, Niki and Markham, Andrew},
  booktitle={Proceedings of the IEEE/CVF conference on computer vision and pattern recognition},
  pages={4977--4987},
  year={2021}
}

@inproceedings{qi2017pointnet,
  title={Pointnet: Deep learning on point sets for 3d classification and segmentation},
  author={Qi, Charles R and Su, Hao and Mo, Kaichun and Guibas, Leonidas J},
  booktitle={Proceedings of the IEEE conference on computer vision and pattern recognition},
  pages={652--660},
  year={2017}
}

@article{qi2017pointnet++,
  title={Pointnet++: Deep hierarchical feature learning on point sets in a metric space},
  author={Qi, Charles Ruizhongtai and Yi, Li and Su, Hao and Guibas, Leonidas J},
  journal={Advances in neural information processing systems},
  volume={30},
  year={2017}
}

@article{wang2019dynamic,
  title={Dynamic graph cnn for learning on point clouds},
  author={Wang, Yue and Sun, Yongbin and Liu, Ziwei and Sarma, Sanjay E and Bronstein, Michael M and Solomon, Justin M},
  journal={ACM Transactions on Graphics (tog)},
  volume={38},
  number={5},
  pages={1--12},
  year={2019},
  publisher={Acm New York, NY, USA}
}

@article{li2018pointcnn,
  title={Pointcnn: Convolution on x-transformed points},
  author={Li, Yangyan and Bu, Rui and Sun, Mingchao and Wu, Wei and Di, Xinhan and Chen, Baoquan},
  journal={Advances in neural information processing systems},
  volume={31},
  year={2018}
}

@inproceedings{graham20183d,
  title={3d semantic segmentation with submanifold sparse convolutional networks},
  author={Graham, Benjamin and Engelcke, Martin and Van Der Maaten, Laurens},
  booktitle={Proceedings of the IEEE conference on computer vision and pattern recognition},
  pages={9224--9232},
  year={2018}
}

@article{armeni2017joint,
  title={Joint 2d-3d-semantic data for indoor scene understanding},
  author={Armeni, Iro and Sax, Sasha and Zamir, Amir R and Savarese, Silvio},
  journal={arXiv preprint arXiv:1702.01105},
  year={2017}
}

@inproceedings{dai2017scannet,
  title={Scannet: Richly-annotated 3d reconstructions of indoor scenes},
  author={Dai, Angela and Chang, Angel X and Savva, Manolis and Halber, Maciej and Funkhouser, Thomas and Nie{\ss}ner, Matthias},
  booktitle={Proceedings of the IEEE conference on computer vision and pattern recognition},
  pages={5828--5839},
  year={2017}
}

@inproceedings{caesar2020nuscenes,
  title={nuscenes: A multimodal dataset for autonomous driving},
  author={Caesar, Holger and Bankiti, Varun and Lang, Alex H and Vora, Sourabh and Liong, Venice Erin and Xu, Qiang and Krishnan, Anush and Pan, Yu and Baldan, Giancarlo and Beijbom, Oscar},
  booktitle={Proceedings of the IEEE/CVF conference on computer vision and pattern recognition},
  pages={11621--11631},
  year={2020}
}

@article{ravi2024sam,
  title={Sam 2: Segment anything in images and videos},
  author={Ravi, Nikhila and Gabeur, Valentin and Hu, Yuan-Ting and Hu, Ronghang and Ryali, Chaitanya and Ma, Tengyu and Khedr, Haitham and R{\"a}dle, Roman and Rolland, Chloe and Gustafson, Laura and others},
  journal={arXiv preprint arXiv:2408.00714},
  year={2024}
}

@inproceedings{xu2021self,
  title={Self-supervised multi-view stereo via effective co-segmentation and data-augmentation},
  author={Xu, Hongbin and Zhou, Zhipeng and Qiao, Yu and Kang, Wenxiong and Wu, Qiuxia},
  booktitle={Proceedings of the AAAI Conference on Artificial Intelligence},
  volume={35},
  number={4},
  pages={3030--3038},
  year={2021}
}

@inproceedings{xu2021digging,
  title={Digging into uncertainty in self-supervised multi-view stereo},
  author={Xu, Hongbin and Zhou, Zhipeng and Wang, Yali and Kang, Wenxiong and Sun, Baigui and Li, Hao and Qiao, Yu},
  booktitle={Proceedings of the IEEE/CVF international conference on computer vision},
  pages={6078--6087},
  year={2021}
}

@article{chen2023costformer,
  title={CostFormer: Cost transformer for cost aggregation in multi-view stereo},
  author={Chen, Weitao and Xu, Hongbin and Zhou, Zhipeng and Liu, Yang and Sun, Baigui and Kang, Wenxiong and Xie, Xuansong},
  journal={arXiv preprint arXiv:2305.10320},
  year={2023}
}

@inproceedings{xu2023semi,
  title={Semi-supervised deep multi-view stereo},
  author={Xu, Hongbin and Chen, Weitao and Liu, Yang and Zhou, Zhipeng and Xiao, Haihong and Sun, Baigui and Xie, Xuansong and Kang, Wenxiong},
  booktitle={Proceedings of the 31st ACM International Conference on Multimedia},
  pages={4616--4625},
  year={2023}
}

@article{xu2024controlrm,
  title={ControLRM: Fast and controllable 3D generation via large reconstruction model},
  author={Xu, Hongbin and Chen, Weitao and Zhou, Zhipeng and Xiao, Feng and Sun, Baigui and Shou, Mike Zheng and Kang, Wenxiong},
  journal={arXiv preprint arXiv:2410.09592},
  year={2024}
}

@inproceedings{xu2026cyc3d,
  title={Cyc3d: Fine-grained controllable 3d generation via cycle consistency regularization},
  author={Xu, Hongbin and Yu, Chaohui and Xiao, Feng and Xing, Jiazheng and Ci, Hai and Chen, Weitao and Wang, Fan and Li, Ming},
  booktitle={Proceedings of the AAAI Conference on Artificial Intelligence},
  volume={40},
  number={21},
  pages={17895--17903},
  year={2026}
}

@article{xu2024robustmvs,
  title={Robustmvs: Single domain generalized deep multi-view stereo},
  author={Xu, Hongbin and Chen, Weitao and Sun, Baigui and Xie, Xuansong and Kang, Wenxiong},
  journal={IEEE Transactions on Circuits and Systems for Video Technology},
  volume={34},
  number={10},
  pages={9181--9194},
  year={2024},
  publisher={IEEE}
}
